\documentclass{IEEEoj}

\usepackage{cite}
\usepackage{amsmath,amssymb,amsfonts,bm}
\usepackage{graphicx,color}
\usepackage{textcomp}
\usepackage{orcidlink}
\usepackage{float}
\usepackage[ruled,vlined]{algorithm2e}
\usepackage{algpseudocode}
\usepackage[stable]{footmisc}
\usepackage[english]{babel}
\usepackage{hyperref}
\usepackage{booktabs}
\usepackage{cite}
\usepackage{comment}
\usepackage{siunitx}

\usepackage{booktabs}
\usepackage{tabularx}
\usepackage{array}
\usepackage[caption=false,font=normalsize,labelfont=sf,textfont=sf]{subfig}
\usepackage{textcomp}
\usepackage{stfloats}
\usepackage{url}
\usepackage{verbatim}
\usepackage{balance}
\usepackage{etoolbox}

\definecolor{TUMblue}{RGB}{0, 101, 189}

\usepackage[acronym, toc=false, shortcuts, nomain, nonumberlist]{glossaries}
\newacronym{GP}{GP}{Gaussian Process}
\newacronym{LTV}{LTV}{linear time-variant}
\newacronym{GPMP2}{GPMP2}{Gaussian  Process  Motion  Planning}
\newacronym{iGPMP2}{iGPMP2}{Incremental Gaussian  Process  Motion  Planning}
\newacronym{MAP}{MAP}{Maximum a Posteriori}
\newacronym{LTV-SDE}{LTV-SDE}{linear  time  variant,  stochastic differential equation}
\newacronym{PDF}{PDF}{ probability density function}
\newacronym{MS2MP}{MS2MP}{Min-sum  Message Passing  algorithm  for  Motion  Planning}
\newacronym{DOF}{DOF}{degree-of-freedom}
\newacronym{CoC}{CoC}{center-of-curvature}
\newacronym{PAI}{PAI}{planning-as-inference}
\newacronym{MDPs}{MDPs}{Markov decision processes}
\newacronym{SLAM}{SLAM}{Simultaneous Localization and Mapping}
\newacronym{PGM}{PGM}{Probabilistic Graphical Model}
\newacronym{GaBP}{GaBP}{Gaussian Belief Propagation}
\newacronym{cGaBP}{cGaBP}{Constrained Gaussian Belief Propagation}
\newacronym{iGaBP}{iGaBP}{Incremental Gaussian Belief Propagation}
\newacronym{GTSAM}{GTSAM}{Georgia Tech Smoothing and Mapping}
\newacronym{DMBP}{DMPB}{Difference-map Belief Propagation}

\newacronym{DC}{DC}{Divide and Concur}
\newacronym{DM}{DM}{Difference Map}
\newacronym{BP}{BP}{Belief Propagation}
\newacronym{SDF}{SDF}{Signed Distance Field}
\newacronym{MPC}{MPC}{Model Predictive Control}
\newacronym{MPCC}{MPCC}{Model Predictive Contouring Control}
\newacronym{N-T}{N-T}{Normal-Tangential}
\newacronym{QP}{QP}{Quadratic Programming}
\newacronym{CoG}{CoG}{center of gravity}
\newacronym{HPIPM}{HPIPM}{High-Performance Interior-Point Method}

\newacronym{WEEE}{WEEE}{Waste Electrical and Electronic Equipment}

\makeglossaries

\SetEndCharOfAlgoLine{}
\SetKwComment{Comment}{}{}

\SetCommentSty{mycommfont}

\newtheorem{assumption}{\textbf{Assumption}}

\newtheorem{remark}{\textbf{Remark}}
\newtheorem{proposition}{\textbf{Proposition}}
\newtheorem{proof}{\textbf{Proof}}

\newtheorem{problem}{\textbf{Problem}}

\def\BibTeX{{\rm B\kern-.05em{\sc i\kern-.025em b}\kern-.08em
    T\kern-.1667em\lower.7ex\hbox{E}\kern-.125emX}}
\AtBeginDocument{\definecolor{ojcolor}{cmyk}{0.93,0.59,0.15,0.02}}
\def\OJlogo{\vspace{-14pt}\includegraphics[height=28pt]{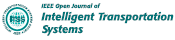}}
\begin{document}
\receiveddate{XX Month, XXXX}
\reviseddate{XX Month, XXXX}
\title{Factor Graph-Based Planning as Inference for Autonomous Vehicle Racing}

\author{Salman Bari\authorrefmark{1}, Xiagong Wang\authorrefmark{1}, Ahmad Schoha Haidari\authorrefmark{1,2}, Dirk Wollherr\authorrefmark{1}}
\affil{TUM School of Computation, Information and Technology, Technical University of Munich, Germany}
\affil{ARRK Engineering GmbH, Munich, Germany}
\corresp{CORRESPONDING AUTHOR: Salman Bari (e-mail: s.bari@tum.de).}
\markboth{Factor Graph-Based Planning as Inference for Autonomous Vehicle Racing}{Bari \textit{et al.}}
\begin{abstract}
Factor graph, as a bipartite graphical model, offers a structured representation by revealing local connections among graph nodes.
This study explores the utilization of factor graphs in modeling the autonomous racecar planning problem, presenting an alternate perspective to the traditional optimization-based formulation.
We model the planning problem as a probabilistic inference over a factor graph, with factor nodes capturing the joint distribution of motion objectives.
By leveraging the duality between optimization and inference, a fast solution to the maximum a posteriori estimation of the factor graph is obtained via least-squares optimization. 
The localized design thinking inherent in this formulation ensures that motion objectives depend on a small subset of variables.
We exploit the locality feature of the factor graph structure to integrate the minimum curvature path and local planning computations into a unified algorithm. 
This diverges from the conventional separation of global and local planning modules, where curvature minimization occurs at the global level.
The evaluation of the proposed framework demonstrated superior performance for cumulative curvature and average speed across the racetrack.
Furthermore, the results highlight the computational efficiency of our approach.
While acknowledging the structural design advantages and computational efficiency of the proposed methodology, we also address its limitations and outline potential directions for future research.
\end{abstract}
\begin{IEEEkeywords}
Motion planning, autonomous racing vehicles, probabilistic inference, factor graph, model predictive control.
\end{IEEEkeywords}


\maketitle
\glsresetall
\section{INTRODUCTION} \label{sec:intro}
\vspace{-16.4cm}
\mbox{\small This work has been accepted for publication in IEEE Open Journal of Intelligent Transportation Systems.
This is the author's version and is subject to change.}
\vspace{14.2cm}
\vspace{-14.7cm}
\mbox{\small ~~ DOI:~https://doi.org/10.1109/OJITS.2024.3418956.}
\vspace{16cm} 
\IEEEPARstart{T}{he} primary objective of an autonomous racing vehicle is to navigate the racetrack with maximum velocity, aiming for the shortest lap time.
Achieving this goal poses a substantial challenge, requiring advanced motion planning algorithms capable of managing the vehicle near its stability limits.
Given the high-speed nature of the movement of the vehicle, the computing efficiency of the planning algorithm emerges as a paramount consideration.
In essence, the planning algorithm must not only address planning objectives but also prioritize computational efficiency. 
Motion planning in autonomous racing is typically divided into three tiers: global planning, local planning, and behavior planning, as noted by Betz et al.~\cite{Betz22}.
Behavioral planning entails details about the racecar's high-level mission planning.
However, for the scope of this work, we will not delve into behavior planning. 
Our focus remains on global planning and local planning aspects.
Global planning focuses on determining the optimal path, known as the raceline~\cite{Heilmeier20},  allowing the racecar to attain higher cornering speeds.
On the other hand, local planning generates a trajectory for a shorter horizon, contingent on the optimal raceline derived from the higher-level global planning module.
The local planning ensures a reliable trajectory that adheres to feasibility criteria, including but not limited to obstacle avoidance, physical limits, and compliance with modeling constraints.

Trajectory optimization emerges as the preferred methodology~\cite{Betz22} for the development of fast and reliable global and local planning algorithms.
However, it is noteworthy that the optimizing criteria differ between the global and local planning modules. 
Global planning typically seeks to optimize for the lowest lap time.
While a minimum time optimization approach theoretically yields an optimal raceline that minimizes lap time, it is computationally expensive. 
Alternatively, geometrical optimization can be employed to generate a raceline with minimum curvature.
Heilmeier et al.~\cite{Heilmeier20} emphasize the preference for this approach due to its relatively lower computational cost, yet it achieves lap times comparable to those obtained through the minimum time optimization methodology.
The optimal raceline entails a balance between the shortest path, which minimizes distance, and the minimum curvature path, which offers higher speeds at turns, leading to an overall faster lap time.

Local planning strategies based on trajectory optimization frame racecar planning as an optimal control problem, facilitating the systematic incorporation of obstacles and constraints.
The~\ac{MPCC} approach proposed in~\cite{Liniger15} stands out as a computationally efficient solution tailored for 1:43 scaled RC cars in which convex optimization using~\ac{QP} is employed to derive the best solution.
Achieving the optimal trajectory involves weighing the performance of following a reference path against maximizing progress.

Inspired by~\cite{Liniger15}, we adopt a similar receding horizon strategy for generating the output trajectory.
However, the planning problem is formulated in the context of \ac{PAI}.
The optimization objective for trajectory planning in an autonomous racecar involves multiple terms that exhibit a local nature, meaning they depend on a limited subset of variables.  
Leveraging this observation, we propose to model the planning problem using a factor graph, which systematically captures this locality structure.
This approach structures all the planning objectives as probability distribution~\cite{Toussaint09, MukadamDYDB18}, represented as factors on the factor graph.
This Bayesian perspective on planning, rooted in trajectory optimization as probabilistic inference \cite{Attias03, Toussaint09 ,ToussaintG10}, has not been explored in the context of autonomous racecar planning to the best of our knowledge.
Our research seeks to fill this gap by framing racecar planning as a probabilistic inference problem on a factor graph.

In our proposed approach, we aim to balance adherence to the reference centerline against progress maximization.
This is achieved by generating a trajectory that minimizes cumulative curvature, thereby enhancing speed. 
By structuring planning objectives as factor nodes on a factor graph, we consolidate planning into a unified framework, systematically integrating motion models, constraints, and curvature minimization to optimize the trajectory for maximizing progress within the planning horizon.
The signed distance field is computed offline from the racetrack boundaries, and then the output trajectory is generated online in a receding horizon manner by estimating \ac{MAP} inference of the factor graph.

While our planning algorithm shares the receding horizon controller concept with~\cite{Liniger15} to maximize racecar progress within a horizon, we innovate by representing optimization objectives as a joint distribution over coupled random variables on a factor graph. Unlike traditional methods that segregate path curvature minimization between global and local planning, our approach integrates curvature minimization into the local planning framework.

We compute the \ac{MAP} estimation of the formulated factor graph via numerical optimization~\cite{DellaertK06} that results in enhanced computational efficiency. 
In order to validate the effectiveness of our proposed framework, we perform a comparative analysis against the \ac{QP} optimization-based \ac{MPCC}~\cite{Liniger15} for two exemplary racetracks. 
The results illustrate the superior performance of our method in terms of both cumulative curvature and average speed, resulting in better lap time.
In summary, the contributions of this work are as follows:
\begin{itemize}
    \item A factor graph-based racecar planning framework is introduced that offers a structured representation of planning objectives highlighting the insights into the local design thinking of modeling the problem. 
    \item We propose a novel methodology to minimize the cumulative curvature of the path by introducing a factor node attached to three consecutive variable nodes, which is integrated into the rest of the factor graph. This leads to a unified planning framework considering raceline curvature, vehicle motion model, and other constraints within a single planning module.
\end{itemize}

The rest of the paper is structured as follows: In Section~\ref{sec:relatedWork}, we provide an overview of the related work in the field. 
Preliminaries, vehicle model adopted in this work, and the background about \ac{PAI} is presented in Section~\ref{sec:prelimAndBackground}.
The problem statement and key methodology of the proposed approach are introduced in Section~\ref{sec:methodology}. 
Section~\ref{sec:FGPlanning} details the formulation of factors based on the motion objectives and constraints. 
Implementation details of the proposed algorithmic framework and the results are presented in Section~\ref{sec:results}. 
A detailed discussion about the limitations of the proposed methodology and future research directions are presented in Section~\ref{sec:discussion}. 
Conclusive remarks are outlined in Section~\ref{sec:conclusions}.
\section{RELATED WORK} \label{sec:relatedWork}
In the context of autonomous racing, research in the field of global planning is very broad, and it can be categorized based on the overall optimization criteria~\cite{Betz22}. 
These optimizing criteria may include lap time~\cite{HerrmannPBL20}, energy consumption~\cite{HerrmannCBL19}, and geometric properties of the raceline~\cite{Braghin08}. 
These diverse optimization goals have led to the formulation of distinct global planning strategies.
Notably, geometrical optimization approaches~\cite{Heilmeier20, Braghin08, kapania16} have gained preference for their computational efficiency in generating an optimal raceline.
Finding an optimal raceline is a complex task, as the shortest path does not necessarily offer the minimum lap time for curved racetracks since the speed is considerably reduced at turns. 
Due to racetrack curves, the path with the minimum curvature offers higher speed.
Therefore, the optimal raceline is a compromise between the shortest path, which minimizes the path length, and the minimum curvature path, which offers higher speed at turns, ultimately resulting in reduced lap time.

Braghin et al.~\cite{Braghin08} proposed a race car driver model that uses geometric optimization of the path along a racetrack to find the minimum curvature raceline. 
The output of the proposed algorithm is the best compromise between the shortest path and path with the minimum curvature, resulting in a speed profile that allows to minimize the lap time.
Kapania et al. ~\cite{kapania16} followed a similar approach and divided the trajectory generation into two sequential steps. 
The first step involved generating a velocity profile, followed by minimizing the path curvature using convex optimization.
Recently,  Heilmeier et al.~\cite {Heilmeier20} proposed a \ac{QP}-based optimization formulation for finding the minimum curvature raceline. 
It has also been noted that the raceline produced by ~\ac{QP} optimization by minimizing the cumulative curvature of the path produces comparable lap time as compared to the minimum time optimization strategy while being computationally efficient.
However, to refine the raceline further, an iterative \ac{QP} optimization strategy is employed, leading to increased computation time.

Similar to existing methods for minimum curvature raceline planning~\cite{Heilmeier20, Braghin08}, our objective is also to minimize the cumulative quadratic curvature of the centerline. 
However, in contrast to conventional \ac{QP}-based optimization approach~\cite{Heilmeier20}, we propose a novel method that utilizes probabilistic inference and factor graph.
We transform the representation of the racetrack and centerline into the \ac{N-T} coordinate system, which leads us to the proposition that the cumulative curvature of an arc can be reduced by minimizing the tangential rotational angle between two waypoints along the centerline.
Based on this, we design factors that consider the minimization of the tangential rotational angle and integrate these into the factor graph that also includes the local planning module related factors.

Regarding local planning, the \ac{MPC} approach has garnered significant attention within the autonomous racing research community in recent years~\cite{Betz22}.
In \ac{MPC} case, the global plan (raceline) is optimized, keeping the consideration of obstacle avoidance, constraints, and the system model.
The cost function is used to represent planning objectives such as progress along the racetrack, obstacle avoidance and deviation from the optimal raceline.
There is a wide range of ~\ac{MPC} approaches proposed in recent years, such as an \ac{MPC} methodology that mimics a professional driver~\cite{AndersonAW16}, introducing a two-mode switching approach aimed at optimizing either for the minimum-time or maximum velocity objective. 
A sampling-based \ac{MPC}~\cite{WilliamsDGRT16} relying on path integral control to achieve entropy minimization. An approach that considers vehicle stabilization, path tracking, and collision avoidance is proposed in~\cite{FunkeBEG17}. Similarly, a nonlinear-\ac{MPC} in~\cite{kalaria21},  and a Stochastic~\ac{MPC} in~\cite{Tim21} to plan efficient overtaking maneuvers by utilizing \ac{GP} to predict the opponent’s maneuver.
Recently, a model-based learning of control policy has been proposed in~\cite{1spielberg23} that updates the control policy based on lap time gradient with respect to control parameters, leading to performance improvement.
However, in terms of computing efficiency, \ac{MPCC} approach~\cite{Liniger15} stands out as a framework that utilizes linear time-varying models at each sampling time to construct local approximations in the form of convex \ac{QP}.
The resultant ~\ac{QP}s are efficiently solved to achieve real-time working of the control scheme for 1:43 scaled RC racecars.

On the other hand, \ac{PAI} emerges as an alternative approach, providing fast solutions to planning problems~\cite{Toussaint09, MukadamDYDB18}, thereby indicating its potential utility in the field of autonomous racing. 
A recent work by Bazzana et al.\cite{Bazzana22} also proposes a factor graph-based \ac{MPC} for navigation. While our algorithm shares similarities with~\cite{Bazzana22} in formulating the planning problem as factor graph-based \ac{MPC}, there are two key distinctions. Firstly, we address the more complex problem of autonomous racing, including the incorporation of curvature minimization alongside navigation. Secondly, we employ least square optimization for \ac{MAP} estimation, ensuring faster computation, a crucial factor in the context of autonomous racing.

While the fundamental idea of our proposed planning algorithm is also to maximize the progress of the racecar within the horizon as a performance measure similar to~\cite{Liniger15}, we represent optimizing objectives as a joint distribution over coupled random variables on the factor graph. 
Unlike traditional approaches that separately minimize path curvature at the global level, our approach eliminates the distinction between global and local planning.
The factor responsible for optimizing minimum curvature is integrated along with other local planning factor nodes on the factor graph.
Other existing works also consider global planning and local planning as a unified framework~\cite{OKelly20, Kloeser20}. 
However, these approaches do not incorporate cumulative curvature minimization in their optimizing criteria. 
Moreover, all these methods adopt the traditional trajectory optimization approach, contrasting with our probabilistic inference-based framework. 
Our approach not only demonstrates computational efficiency and superior performance but also introduces a local perspective within the structured representation of motion objectives on a factor graph.

\section{PRELIMINARIES and BACKGROUND}  \label{sec:prelimAndBackground}
The racetrack is defined by left and right track boundaries, denoted as $\bm{b}_{l} \rightarrow \mathbb{R}^{2}$ and $\bm{b}_{r} \rightarrow \mathbb{R}^{2}$, respectively.
The centerline $\bm{p}$ of the racetrack is assumed to be available as a set of discretized waypoints, represented by $\bm{p} = \begin{bmatrix} \bm{p}_{i} & \cdots & \bm{p}_{N} \end{bmatrix}^{\top} \rightarrow \mathbb{R}^{2}$, equidistant from $\bm{b}_{l}$ and $\bm{b}_{r}$.
The workspace of the racing vehicle $\chi \subset \mathbb{R}^{D}$ is partitioned into obstacle-space $\chi_{\mathrm{obs}} \subset \chi$, and the space within the racetrack boundaries $\chi_{\mathrm{track}} \subset \chi$ which is considered free space. 
The racing vehicle's state is denoted as $\bm{\theta}\left ( t \right ) : t\rightarrow \mathbb{R}^{D}$, where $D$ is the state dimensionality, and $\bm{\theta}\left ( t \right )$ is a continuous time function mapping time $t$ to vehicle states $\bm{\theta}$.
Following Mukadam et al.~\cite{MukadamDYDB18}, racing vehicle states are sampled from a continuous time \ac{GP},
\begin{equation}
    \bm{\theta}\left ( t \right ) = \left [ \bm{\theta}_{0} ,\; . . .,\;\bm{\theta}_{N} \right ]^{T} \sim 
    \mathcal{N}\left ( \bm{\mu }\left (t \right ), \bm{\mathcal{K}}\right ), 
    \label{eq:problem:jointDistribution}
\end{equation}
where $N$ represents the total number of states for a set of times $\bm{t}=t_0,\dots, t_N$.
Here, $\bm{\mu }\left ( t \right )$ is the mean vector, and $\bm{\mathcal{K}}\left ( t, {t}' \right ) $ is the covariance function matrix.
The racecar planning problem is characterized as finding the optimal control $\bm{u}^{*}$ that generates optimal state $\bm{\theta}^{*}$. 
\subsection{FACTOR GRAPH FOR PLANNING AS INFERENCE} \label{sec:backgroundDef}
In this section, we revisit the formulation of the planning problem as probabilistic inference following~\cite{ToussaintG10, MukadamDYDB18} on a factor graph. 
In a general planning problem, we are interested in finding the optimal state sequence $\bm{\theta}^{*}_{0:N}$ and control sequence $\bm{u}^{*}_{0:N-1}$ that maximizes the joint probability distribution  $p\left ( \bm{\Theta}\right )$. Where, $\bm{\Theta} = \left \{\bm{\theta}, \bm{u} \right \}$.
To model the belief over continuous, multivariate random variables $\bm{\Theta}$, a factor graph $\mathcal{G} = \left ( \bm{\Theta} , \mathcal{F}, \mathcal{E} \right )$ is employed. The joint probability distribution $p\left ( \bm{\Theta}\right )$ is factorized as follows:
\begin{equation}
p\left ( \bm{\Theta}\right ) \propto \prod_{m=1}^{M} \;f_{m}\left ( \bm{\Theta}_{m} \right ),
\label{eq:factorGraph:factorization}
\end{equation}
where $f_{m} \in \mathcal{F}$ are factors associated with the corresponding variable nodes $\bm{\Theta}_{m} \in \bm{\Theta}$ connected through the edges $\mathcal{E}$ of the factor graph.

The optimal state $\bm{\theta}^{*}_{0:N}$ and control input $\bm{u}^{*}_{0:N-1}$ are computed by obtaining \ac{MAP} inference,
\begin{equation}
    \bm{\Theta}^{*} = \underset{\bm{\Theta} }{\arg\max } \prod_{m=1}^{M} \;f_{m}\left ( \bm{\Theta}_{m} \right ).
    \label{eq:Mapinf}
\end{equation} 
This factor graph formulation~\eqref{eq:Mapinf} offers a structured representation of the planning problem, where system state and control input variables are portrayed as variable nodes.
An inference algorithm on the factor graph is utilized to compute the \ac{MAP} estimation, aligning with the planning objectives dictated by the attached factors.

The optimal control sequence $\bm{u}^{*}$ is determined by minimizing the negative logarithm of the probability distribution $p\left ( \bm{\Theta} \right )$ represented over the factor graph $\mathcal{G}$. 
This results in an unconstrained nonlinear least squares optimization problem~\cite{MukadamDYDB18}, which is a well-explored domain with numerous numerical algorithms available, such as the Gauss-Newton or Levenberg-Marquardt algorithms. 
These algorithms iteratively solve a quadratic approximation until convergence is achieved.
The proposed planning problem formulation for autonomous racing and factors are discussed in detail in Section~\ref{sec:methodology} and~\ref{sec:FGPlanning}, respectively.
\subsection{VEHICLE MODEL} \label{sec:vehModel}
\begin{figure}[t]
    \centering
    \includegraphics[width =\linewidth]{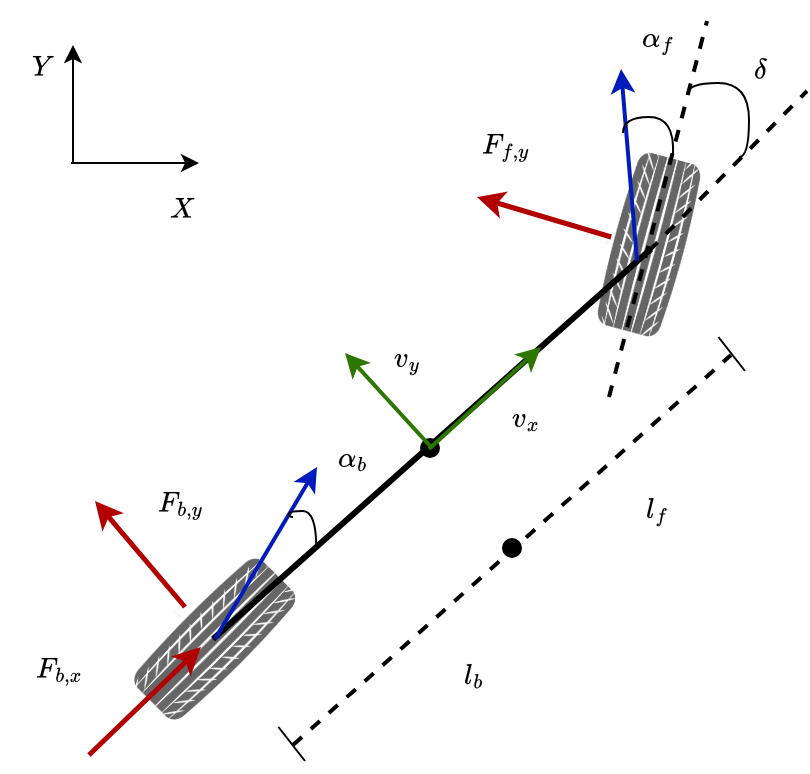}
    \caption{The schematic drawing of the vehicle model with mass $m$. $l_f$ and $l_b$ depict the distance from the \acs{CoG} to the front and rear wheels. The vectors in red illustrate the forces between the tire and the road, while $v_x$ and $v_y$ denote velocity components, and $\alpha_{f}$ and $\alpha_{b}$ represent slip angles.}
    \label{fig:vehicleModel}
\end{figure}

We opt for a bicycle model to represent the dynamics of the racecar, incorporating nonlinear tire force laws.
This model is commonly employed in autonomous racing applications~\cite{Liniger15, Kabzan20}.

The selected vehicle model relies on specific assumptions. These assumptions include the vehicle operating on a flat surface, neglecting load transfer and combined slip, and the assumption that longitudinal drive-train forces act on the \acf{CoG}. 
The vehicle state $ \bm{\theta} = \left [ x, y, v_{x}, v_{y}, \varphi, \omega \right ]^{T}$ consists of the position of the vehicle $\bm{p}=\left [ x, y \right ]^{T}$, the longitudinal and lateral velocity $\bm{v}=\left [ v_{x}, v_{y} \right ]^{T}$, and the rotation component $ \bm{r} = \left [ \varphi, \omega \right ]^{T}$ that includes heading angle $\varphi$ and yaw rate $\omega$.
The control input $\bm{u} = \left [ \delta, d \right ]^{T}$ constitutes the steering angle $\delta$ and the PWM duty cycle $d$ of the electric drive train motor.
The model, represented as $\bm{\dot{\theta}} = \bm{g}\left ( \bm{\theta}, \bm{u}\right )$, is depicted in Fig.~\ref{fig:vehicleModel}.  The associated nonlinear equation of motion is:
\begin{equation}
\bm{\dot{\theta}}
=
\begin{bmatrix}
\begin{aligned}
&v_x\cos  \varphi - v_y\sin  \varphi \\ 
&v_x\sin  \varphi  - v_y\cos \varphi \\ 
&\frac{1}{m}\left ( F_{b,x} - F_{f,y}\sin\delta + mv_y\omega \right )\\ 
&\frac{1}{m}\left ( F_{b,y} - F_{f,y}\cos\delta + mv_x\omega \right )\\ 
&\omega\\ 
&\frac{1}{I_z}\left ( F_{f,y}l_f\cos\delta - F_{b,y}l_b \right )
\end{aligned}
\end{bmatrix}
=:
\bm{g}\left ( \bm{\theta}, \bm{u}\right ), 
    \label{eq:model:motion}
\end{equation}
where the car is characterized by a mass $m$ and inertia $I_z$,  $l_f$ and $l_b$ denote the distances from the \ac{CoG} to the front and rear wheels axle, respectively.
The tire forces $F$ capture the interaction between the car and the road, with subscripts $x$ and $y$ indicating longitudinal and lateral forces, and $f$ and $b$ representing the front and rear wheels.

Given the objective of racing, it is crucial for the tire forces model to realistically depict the racecar's behavior at high speeds and handling limits. To strike a balance between precision and computational efficiency, a simplified Pacejka tire model~\cite{bakker87} has been selected. This choice is informed by the need to ensure a realistic representation of the racecar's dynamics during high-speed racing scenarios.

\begin{subequations}
\begin{equation}
\begin{aligned}
& F_{f,y} = D_f \sin \left ( C_f \arctan \left ( B_f \alpha_f  \right ) \right ), \; \\
\mathrm{s.t.}\; \;
&  \alpha_f = -\arctan \left ( \frac{\omega l_f + v_y}{v_x} \right) + \delta,
    \label{eq:model:tyreModelFrontY}
\end{aligned}
\end{equation}
\begin{equation}
\begin{aligned}
& F_{b,y}  = D_b \sin \left ( C_b \arctan \left ( B_b \alpha_b  \right ) \right ), \\
\mathrm{s.t.}\; \;
& \alpha_b = -\arctan \left ( \frac{\omega l_b - v_y}{v_x} \right),
    \label{eq:model:tyreModelBackY}
\end{aligned}
\end{equation}
\begin{equation}
\begin{aligned}
& F_{b,x} = \left (C_{m1} - C_{m2} v_x \right )d - C_b - C_d v_{x}^{2}.
    \label{eq:model:tyreModelBackX}
\end{aligned}
\end{equation}
\end{subequations}
The parameters $B$, $C$ and $D$  are experimentally identified to define the shape of the semi-empirical curve.
Whereas, $\alpha$ is the slip angle.
The longitudinal force of the rear wheel, $F_{b,x}$~\eqref{eq:model:tyreModelBackX}, is modeled employing a motor model for the DC electric motor.
We linearize the model~\eqref{eq:model:motion} with respect to the current state, followed by discretization step.
The resulting discretized model is utilized to predict future states at steps $j = 0,...,n$, where $n$ is the prediction horizon. 
This prediction is based on the current state $\bm{\theta}_0$ and the input sequence over the prediction horizon, $\bm{u}_j \; , \; j = 0,...,n-1$.
Refer to (cf.~\cite{Liniger15}, Sec. II) for more details regarding the racecar model.

\section{PROBLEM STATEMENT AND APPROACH} \label{sec:methodology}
This work addresses the challenge of determining the optimal trajectory for an autonomous racecar navigating a racetrack. The objective is to identify the optimal control input $\bm{u}^{*}$ that produces the optimal trajectory. This optimal trajectory seeks to maximize progress while simultaneously adhering to track boundaries and generating a path with minimal curvature, thereby offering a better velocity profile. Formally, the problem addressed by this work is defined as:

\begin{problem} \label{pr:racingPlanning}
Given a racetrack with boundaries $\bm{b}_{l}$ and $\bm{b}_{r}$ and centerline $\bm{p}$, find the optimal control input $\bm{u}^{*}$ that results in optimal state $\bm{\theta}^{*}$ from the start position $\bm{p}_{0}$ to the goal position $\bm{p}_{N}$ by taking into account of generating a minimum curvature path resulting from optimizing the global summed quadratic curvature $k$. The output trajectory should remain within the racetrack boundaries, and both the state and control input must adhere to the constraints specified below,
\begin{subequations}
\begin{equation}
\begin{aligned}
& \bm{p} \in \chi_{\mathrm{track}},
    \label{eq:prob:obstacle}
\end{aligned}
\end{equation}
\begin{equation}
\begin{aligned}
& \bm{r}_{\mathrm{min}} \le \bm{r} \le \bm{r}_{\mathrm{max}}, 
\label{eq:prob:state}
\end{aligned}
\end{equation}
\begin{equation}
\begin{aligned}
& \bm{u}_{\mathrm{min}} \le \bm{u} \le \bm{u}_{\mathrm{max}}.
    \label{eq:prob:control}
\end{aligned}
\end{equation}
\end{subequations}
Here, $\bm{r} = \left [ \varphi, \omega \right ]^{T}$ captures the constraints as $\bm{r}_{\mathrm{min}}$ and $\bm{r}_{\mathrm{max}}$ on the heading angle $\varphi$ and yaw rate $\omega$. 
\end{problem}

\subsection{METHODOLOGY}
To address the Problem~\ref{pr:racingPlanning}, we find the optimal control and state, denoted as $\bm{\Theta}^{*}$, that maximizes the conditional posterior $p\left ( \bm{\Theta} | \mathbf{e}\right )$. 
The prior on the trajectory is represented as $p\left ( \bm{\Theta} \right )$, encompassing the initial belief regarding $\bm{\theta}$ and $\bm{u}$.
Similarly, $l\left ( \bm{\Theta};\mathbf{e} \right )$ is the likelihood of $\bm{\Theta}$ given conditional events $\mathbf{e}$ on $\bm{\Theta}$. 
Given the prior $p\left ( \bm{\Theta} \right )$ and likelihood $l\left ( \bm{\Theta};\mathbf{e} \right )$, the optimal $\bm{\Theta}^{*}$ is obtained by \ac{MAP} inference,
\begin{equation}
\begin{aligned}
     \bm{\Theta}^{*} & = \underset{\bm{\Theta} }{\arg\max} \;
            \underbrace{ p\left ( \bm{\Theta} | \mathbf{e}\right )}_{ p\left ( \bm{\Theta} \right )l\left ( \bm{\Theta};\mathbf{e} \right )}.
    \label{eq:planningProblem}
\end{aligned}
\end{equation}
The posterior distribution $p\left ( \bm{\Theta} | \mathbf{e}\right )$ in~\eqref{eq:planningProblem} can be represented on a factor graph~\cite{MukadamDYDB18,Toussaint09} by factorizing the prior $ p\left ( \bm{\Theta} \right )$ and likelihood $l\left ( \bm{\Theta};\mathbf{e} \right )$. 
The prior $ p\left ( \bm{\Theta} \right )$ is factorized as,
\begin{equation}
\begin{aligned}
      p\left ( \bm{\Theta} \right ) \propto & \; f_{0}^{\mathrm{s}}\left ( \bm{p}_{0} \right ) f_{N}^{\mathrm{g}}\left ( \bm{p}_{N} \right )
 \prod_{i=0}^{N}f_{i}^{\mathrm{ref}}\left ( \bm{p}_{i} \right ) \\
        & \prod_{i=0}^{N}f_{i}^{\mathrm{vel}}\left ( \bm{v}_{i} \right ) \prod_{i=0}^{N}f_{i}^{\mathrm{rlim}}\left ( \bm{r}_{i} \right ) 
        \prod_{i=0}^{N-1}f_{i}^{\mathrm{ulim}}\left ( \bm{u}_{i} \right ),
    \label{eq:factorsPrior}
\end{aligned}
\end{equation}
where the factors $f_{0}^{\mathrm{s}}\left ( \bm{p}_{0} \right )$ and $ f_{N}^{\mathrm{g}}\left ( \bm{p}_{N} \right )$ are used to provide the functionality of fixing the start position and goal position. 
The prior probability distribution of centerline and desired velocity is captured by the factors $f_{i}^{\mathrm{ref}}$ and $f_{i}^{\mathrm{vel}}$ respectively. 
The constraints for $\bm{r}$ from~\eqref{eq:prob:state}, and for $\bm{u}$ from~\eqref{eq:prob:control} are imposed as soft constraints by the factors $f_{i}^{\mathrm{rlim}}$ and $f_{i}^{\mathrm{ulim}}$, respectively.
Similarly, we factorize the likelihood $l\left ( \bm{\Theta};\mathbf{e} \right )$ as,
\begin{equation}
\begin{aligned}
      l\left ( \bm{\Theta};\mathbf{e} \right ) \propto & \; 
                        \prod_{i=1}^{N-1}f_{i}^{\mathrm{obs}}\left ( \bm{p}_{i} \right ) 
                        \prod_{i=1}^{N}f_{i}^{\mathrm{sys}}\left ( \bm{\theta}_{i-1}, \bm{u}_{i-1}, \bm{\theta}_{i} \right ) \\
        & \prod_{i=0}^{N-2}f_{i}^{\mathrm{curv}}\left ( \bm{p}_{i}, \bm{p}_{i+1}, \bm{p}_{i+2} \right ).
    \label{eq:factorsLikelihood}
\end{aligned}
\end{equation}
The likelihood function incorporates obstacle avoidance via the factor $f_{i}^{\mathrm{obs}}$. 
This ensures that the output trajectory remains inside the boundary limits as per the requirement in~\eqref{eq:prob:obstacle}. 
The system dynamics are included via the factor $f_{i}^{\mathrm{sys}}$, and $f_{i}^{\mathrm{curv}}$ is used for curvature minimization.
The detailed formulation of the factors is discussed in Section~\ref{sec:FGPlanning}.
\begin{figure}[t]
    \centering
    \includegraphics[width =\linewidth]{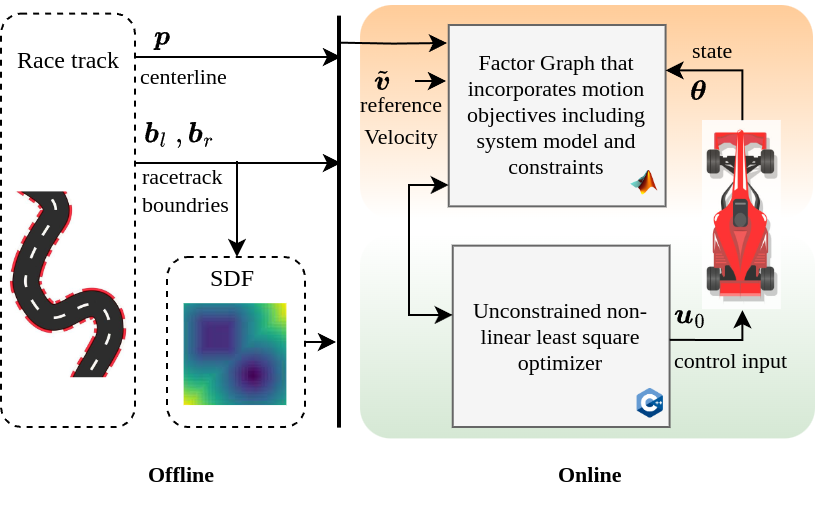}
    \caption{Overall methodology of the proposed framework. The offline phase includes computations conducted prior to the initiation of the race. The online phase involves trajectory generation.}
    \label{fig:methodology}
\end{figure}
\begin{remark}  \label{re:refVeloc}
The factor $f_{i}^{\mathrm{vel}}$ corresponds to monitoring the prior belief regarding velocity, denoted as $p\left ( \bm{v} \right )$. This factor serves as a reference for the maximum achievable velocity $\Tilde{\bm{v}}$, which is assumed to be known.\newline
Also, $\bm{p}_{N}$ represents the racing finish line, and the racecar does not stop at this point, making it close to the real-world racing scenario. Instead, it is assumed that the planning extends to $N+M$ steps, where $M$ covers the deceleration part in the end. 
\end{remark}

\begin{figure*}[t]
    \centering
    \includegraphics[width =\linewidth]{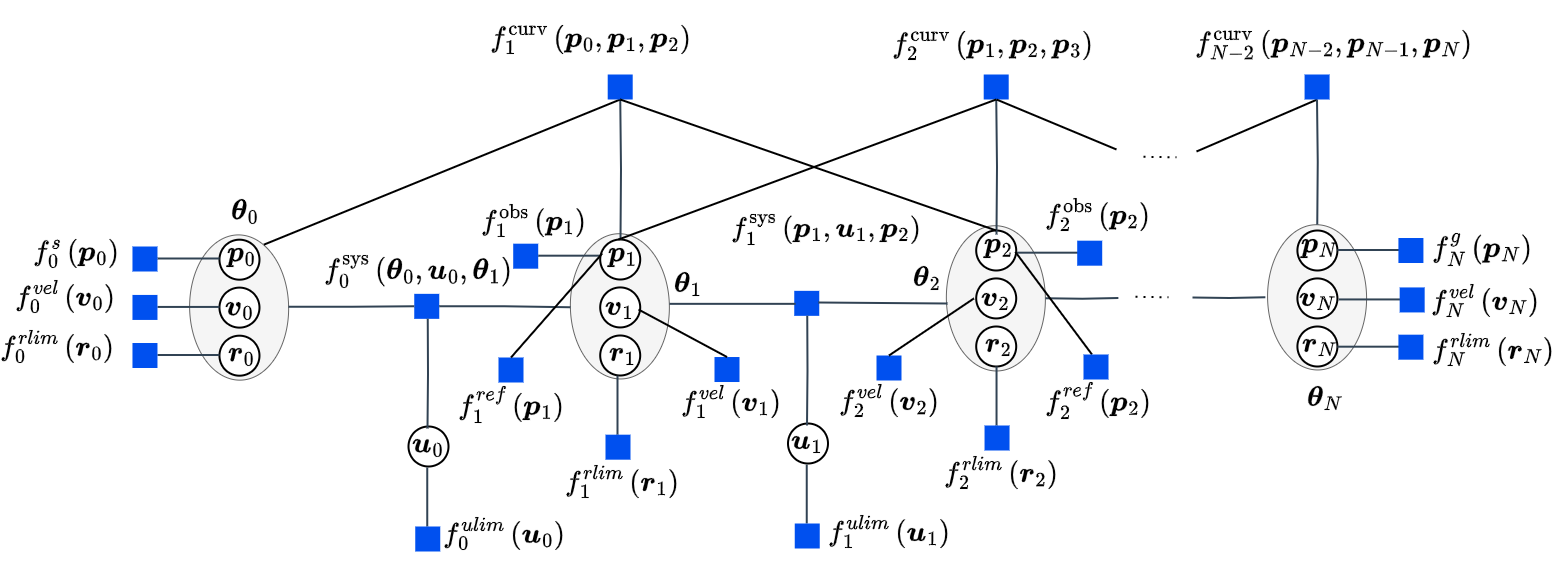}
    \caption{Architecture of the factor graph for autonomous racecar planning. Note that the factor nodes are affixed to the state and control components, presented in the form of variable nodes emphasizing the local connections among them.}
    \label{fig:factorGraphFormulation}
\end{figure*}

The methodology employed in this study is depicted in Fig.~\ref{fig:methodology}. 
It consists of two distinct phases: an offline phase and an online phase. 
During the offline phase, \ac{SDF} is computed based on the racetrack boundaries, which is then used in obstacle avoidance factors (ref. to Sec.~\ref{sec:likelihoodFactors} for details). 
In the online phase, we initiate the process by formulating the factor graph and subsequently solving it for the prediction horizon $n$.
The factor graph formulation of planning problem~\eqref{eq:planningProblem} is illustrated in Fig.~\ref{fig:factorGraphFormulation}, which highlights local connections among graph nodes. 

An inference algorithm, applied to the factor graph, computes the posterior distribution over all trajectories, adhering to planning objectives dictated by the attached factors. 
The efficient solution of the factor graph is achieved through the utilization of sparse least squares. This approach, widely used in simultaneous localization and mapping \cite{DellaertK06} and planning algorithms \cite{MukadamDYDB18}, transforms the sparse graphical model representation into an unconstrained least squares optimization problem.
We convert the \ac{MAP} inference problem into a least square optimization problem by taking the negative logarithm of the posterior distribution. 
Adopting a similar approach as in\cite{MukadamDYDB18}, we employ the Levenberg–Marquardt algorithm to solve the least square optimization problem.
In the following section, we elaborate on the detailed design of the factor nodes.
\section{FORMULATION OF FACTOR NODES} \label{sec:FGPlanning}
The factorization of the prior and likelihood in~\eqref{eq:factorsPrior} and~\eqref{eq:factorsLikelihood} consist of different kinds of factor nodes. 
These nodes are associated with distinct objectives, each defined by error functions based on their connections with variable nodes in the graph. The detailed description of each factor in both the prior and likelihood is provided in the following sections.

\subsection{FACTORIZATION OF PRIOR} \label{sec:priorfactors}
Within our framework, constraints are managed in a soft manner, treating them as prior knowledge on trajectory states and control input. 
The factors $f_{0}^{\mathrm{s}}\left ( \bm{p}_{0} \right )$ and $f_{N}^{\mathrm{g}}\left ( \bm{p}_{N} \right )$ establish the prior distribution for the start and goal positions. 
Following the methodology from~\cite{MukadamDYDB18}, the factor error function is defined as the Mahalanobis distance, enabling the incorporation of the constraint information as,
\begin{equation}
f_{0}^{\mathrm{s}}\left ( \bm{p}_{0} \right ) =  \exp \left \{ -\frac{1}{2}\left \| \bm{e}_{0} 
\right \|^{2}_{{\bm{\mathcal{K}}}_{s}}
 \right \},
    \label{eq:problem:priorStart}
\end{equation}
\begin{equation}
f_{N}^{\mathrm{g}}\left ( \bm{p}_{N} \right ) =  \exp \left \{ -\frac{1}{2}\left \| \bm{e}_{N} \right \|^{2}_{{\bm{\mathcal{K}}}_{g}}
 \right \},
    \label{eq:problem:priorGoal}
\end{equation}
where $\bm{e}_{0} = \bm{p}_{0} -\bm{\mu }_{0}$ and $\bm{e}_{N} = \bm{p}_{N} -\bm{\mu }_{N}$, represents the differences from the known start $\bm{\mu }_{0}$ and goal $\bm{\mu }_{N}$. 
The kernels ${\bm{\mathcal{K}}}_{s}$ and ${\bm{\mathcal{K}}}_{g}$ are covariance matrices introducing soft constraints managed by their values. 
Likewise, the centerline information of the racetrack and the maximum desired velocity information are considered as prior knowledge for $f_{i}^{\mathrm{ref}}\left ( \bm{p}_{i} \right )$ and $f_{i}^{\mathrm{vel}}\left ( \bm{v}_{i} \right )$, respectively.
\begin{equation}
\begin{aligned}
f_{i}^{\mathrm{ref}}\left ( \bm{p}_{i} \right ) & =  \exp \left \{ -\frac{1}{2}\left \| \bm{e}_{i}^{\mathrm{ref}} 
\right \|^{2}_{{\bm{\mathcal{K}}}_{\mathrm{ref}}}
 \right \}, \\
\mathrm{s.t.}\; \; \bm{e}_{i}^{\mathrm{ref}} &= \bm{p}_{i} -\bm{\mu }_{i},
    \label{eq:problem:priorRef}
\end{aligned} 
\end{equation}
similarly,
\begin{equation}
\begin{aligned}
f_{i}^{\mathrm{vel}}\left ( \bm{v}_{i} \right ) & =  \exp \left \{ -\frac{1}{2}\left \| \bm{e}_{i}^{\mathrm{vel}} 
\right \|^{2}_{{\bm{\mathcal{K}}}_{\mathrm{vel}}}
 \right \}, \\
\mathrm{s.t.}\; \;  \bm{e}_{i}^{\mathrm{vel}} &= \bm{v}_{i} - \Tilde{\bm{v}}_{i},
\end{aligned}    
    \label{eq:problem:priorVel}
\end{equation}
where $\Tilde{\bm{v}}$ is the desired maximum velocity. 

We use hinge loss to define the inequality soft constraints as, 
\begin{equation}
\begin{aligned}
f_{i}^{\mathrm{rlim}}\left ( \bm{r}_{i} \right ) & =  \exp \left \{ -\frac{1}{2}\left \| \bm{e}_{i}^{\mathrm{rlim}}\left ( \bm{r}_{i} \right ) 
\right \|^{2}_{{\bm{\mathcal{K}}}_{\mathrm{rlim}}}
 \right \}, \\
\mathrm{s.t.}\; \;
\bm{e}_{i}^{\mathrm{rlim}}\left ( \bm{r}_{i} \right )  &=
\Biggl\{\begin{matrix}
\bm{r} - \bm{r}_{\mathrm{min}} & if \;\bm{r} < \bm{r}_{\mathrm{min}} \\
\bm{r} - \bm{r}_{\mathrm{max}} & if \; \bm{r} > \bm{r}_{\mathrm{max}} \\
0 &  \mathrm{else} 
\end{matrix}
    \label{eq:problem:priorRlim}
\end{aligned} 
\end{equation}
similarly,
\begin{equation}
\begin{aligned}
f_{i}^{\mathrm{ulim}}\left ( \bm{u}_{i} \right ) & =  \exp \left \{ -\frac{1}{2}\left \| \bm{e}_{i}^{\mathrm{ulim}}\left ( \bm{u}_{i} \right ) 
\right \|^{2}_{{\bm{\mathcal{K}}}_{\mathrm{ulim}}}
 \right \}. \\
\mathrm{s.t.}\; \;
\bm{e}_{i}^{\mathrm{ulim}}\left ( \bm{u}_{i} \right )  &=
\Biggl\{\begin{matrix}
\bm{u} - \bm{u}_{\mathrm{min}} & if \;\bm{u} < \bm{u}_{\mathrm{min}} \\
\bm{u} - \bm{u}_{\mathrm{max}} & if \; \bm{u} > \bm{u}_{\mathrm{max}} \\
0 &  \mathrm{else} 
\end{matrix}
    \label{eq:problem:priorUlim}
\end{aligned} 
\end{equation}

\begin{remark}  \label{re:prior}
 The covariance matrices, such as ${{\bm{\mathcal{K}}}_{\mathrm{s}}}$, ${{\bm{\mathcal{K}}}_{\mathrm{g}}}$, ${{\bm{\mathcal{K}}}_{\mathrm{rlim}}}$, and ${{\bm{\mathcal{K}}}_{\mathrm{ulim}}}$ are assigned very small values to enforce tight adherence to the constraints imposed by respective factor nodes.
 On the other hand, the values of ${{\bm{\mathcal{K}}}_{\mathrm{ref}}}$ and ${{\bm{\mathcal{K}}}_{\mathrm{vel}}}$ are selected to allow for looser compliance with these constraints, introducing the desirability.
\end{remark}

\subsection{FACTORIZATION OF LIKELIHOOD}   \label{sec:likelihoodFactors}
\begin{figure}[t]
    \centering
    \includegraphics[width =\linewidth]{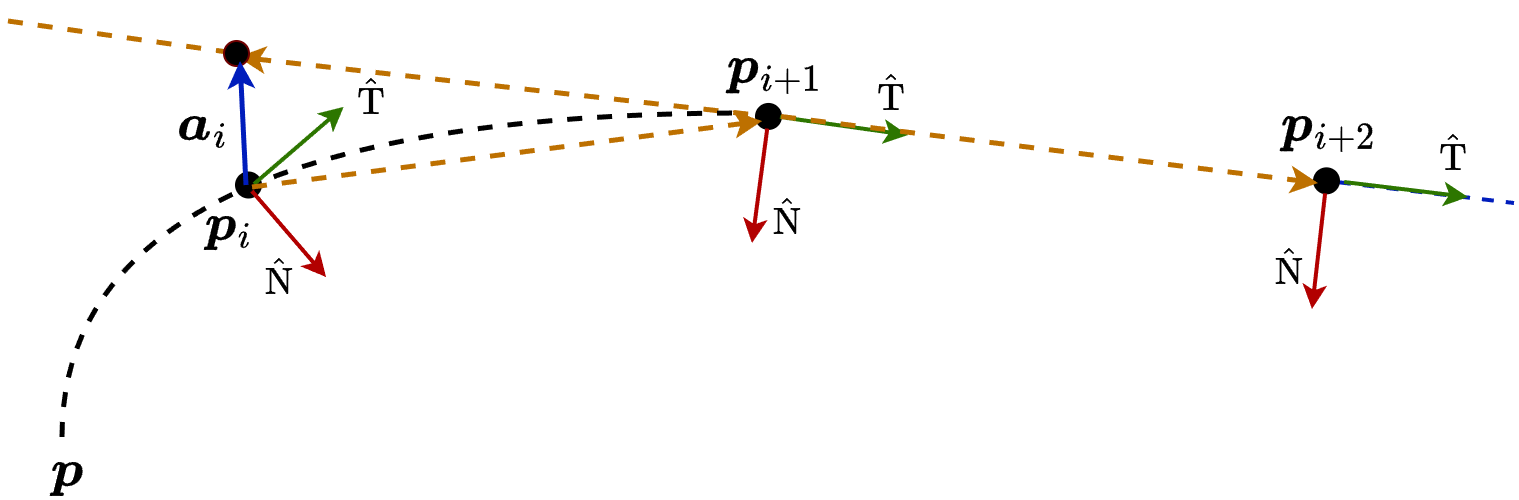}
    \caption{Representation of vectors formed among $\bm{p}_{i}$, $\bm{p}_{i+1}$, and $\bm{p}_{i+2}$ for the minimum curvature factor $f_{i}^{curv}$.}
    \label{fig:curvFactor}
\end{figure}
We devised the likelihood factor to encapsulate the system dynamics by modeling the error as a soft constraint,
\begin{equation}
\begin{aligned}
f_{i}^{\mathrm{sys}}\left ( \bm{\theta}_{i-1}, \bm{u}_{i-1}, \bm{\theta}_{i} \right ) & =  \exp \left \{ -\frac{1}{2}\left \| \bm{e}_{i}^{\mathrm{sys}} 
\right \|^{2}_{{\bm{\Sigma}_{\mathrm{sys}}}}
 \right \}, \\
\mathrm{s.t.}\; \;  \bm{e}_{i}^{\mathrm{sys}} &= \bm{\theta}_{i} - \bm{g}\left (.\right ),
\end{aligned}    
    \label{eq:problem:sysFactor}
\end{equation}
where $\bm{g}\left ( .\right )$ represents the system model as described in~\eqref{eq:model:motion}.
The factor $f^{\mathrm{sys}}$ provides the mapping of state $\bm{\theta}_{i-1}$ and control $\bm{u}_{i-1}$ on to next state $ \bm{\theta}_{i}$. 

The factor $f_{i}^{\mathrm{obs}}\left ( \bm{p}_{i} \right ) $, which specifies the probability of avoiding obstacles is defined as 
\begin{equation}
f_{i}^{\mathrm{obs}}\left ( \bm{p}_{i} \right )  =  \exp \left \{ -\frac{1}{2}\left \| \mathbf{e}_{i}^{\mathrm{obs}} \left (\bm{p}_{i} \right )
\right \|_{\bm{\Sigma }_{\mathrm{obs}}}^{2}
 \right \},
     \label{eq:problem:likelihood}
\end{equation}
where $\mathbf{e}^{\mathrm{obs}} \left (\bm{p} \right )$ represents the vector-valued obstacle cost, and $\bm{\Sigma }_{\mathrm{obs}}$ is a hyperparameter matrix.

In order to design the minimum curvature factor $f_i^{curv}$, we start with the assumption;
\begin{assumption} \label{assum:curv}
Assuming a curvilinear motion of the racecar, the centerline of the racecar $\bm{p} = \begin{bmatrix} \bm{p}_{i} &\cdots  & \bm{p}_{N} \end{bmatrix}^T$ is parameterized by equidistant waypoints and it is represented in an \ac{N-T} coordinate system.
The N-axis is normal to the T-axis and points towards the \ac{CoC}, while the T-axis is tangent to the curve and points in the direction of the racecar's motion. 
\end{assumption}
In an \ac{N-T} coordinate system (from Assumption~\ref{assum:curv}),  the two equidistant waypoints $\bm{p}_{i+1},\bm{p}_{i+2}$ that lie on the same straight line will have unit tangent vector rotation angle $\Delta \Psi = 0$.
Then, the curvature $k_{i}$ is,
\begin{equation}
k_{i} = \lim_{\Delta t\rightarrow \;0} \left |\frac{\Delta \Psi_{i}}{\Delta S_{i}}   \right |,   \;\;\; \forall i, 
    \label{eq:meanCurv}
\end{equation}
where, $\Delta \Psi_{i}$ represents the unit tangent vector rotation angle between $\bm{p}_{i}$ and $\bm{p}_{i+1}$. $\Delta t$ is time and $S_{i}$ represents the arc length.
Note that minimizing the tangential rotation angle $\Delta\Psi$ in~\eqref{eq:meanCurv} results in reduced curvature of the arc $\bm{p}$, which leads us to the following proposition.
\begin{proposition} \label{prop:angle}
For a racecar moving along a non-degenerate arc parameterized by centerline $\bm{p}$, minimizing the tangential rotation angle $\Delta \Psi_{i}$ between two consecutive waypoints $\bm{p}_{i}$ and $\bm{p}_{i+1}$  will result in reduced cumulative curvature $k$. Therefore, the minimum curvature planning objective can be equivalently described as,
\begin{equation}
 \min \sum_{i=0}^{N} k_{i} \cong   \min \sum_{i=0}^{N-1}\Delta \Psi_{i}.
    \label{eq:proposition}
\end{equation}
\end{proposition}
\begin{proof}
See Appendix A.
\end{proof}

In order to get $\Delta\Psi_{i}=0$ for $\bm{p}_{i}$ and $\bm{p}_{i+1}$, we consider $\bm{p}_{i+1}$ as the reference point and adjust the  $\bm{p}_{i}$ so that it lies on the same straight line as $\bm{p}_{i+1}$ and $\bm{p}_{i+2}$ resulting in reduced curvature.
Fig. \ref{fig:curvFactor} represent the vectors formed among the $\bm{p}_{i}$, $\bm{p}_{i+1}$, $\bm{p}_{i+2}$.
Vector $\bm{a}_{i}$ provides the projection of $\bm{p}_{i}$ onto the straight line and it is considered the $f^{curv}$ error function. Hence, 
\begin{equation}
\begin{aligned}
f_{i}^{curv}\left ( \bm{p }_{i}, \bm{p}_{i+1}, \bm{p}_{i+2}\right ) &= 
 \exp \left \{ -\frac{1}{2}\left \|\bm{e}^{\mathrm{curv}}_{i} 
\right \|_{\bm{\Sigma}_{\mathrm{curv}}}^{2}
 \right \}, \\
\mathrm{s.t.}\; \; \bm{e}^{\mathrm{curv}}_{i} &=  \bm{a}_{i}.
    \label{eq:problem:curvFactor}
\end{aligned}
\end{equation}
where $\bm{\Sigma}_{\mathrm{curv}}$ is the hyperparameter matrix. 
The decision to attach additional minimum curvature factors $f_{i}^{curv}$ beyond $N-2$ is optional. 
Typically, attaching factors up to $N-2$ is sufficient to cover all the centerline states. However, in a scenario where the raceline is computed for multiple laps, it is recommended to attach two additional factors up to $N$ to ensure that the raceline smoothly transitions from the last state to the first state of the next lap.
\begin{remark}  \label{re:likelihood}
 The hyperparameter matrix $\bm{\Sigma}_{\mathrm{sys}}$ for factor $f^{\mathrm{sys}}$ is assigned a very small value to tighten the adherence to the system model. 
 Whereas the value of $\bm{\Sigma}_{\mathrm{obs}}$ is selected based on the racetrack, and it usually changes in case of major physical differences in the geometry of the racetrack. $\bm{\Sigma}_{\mathrm{curv}}$ has relatively higher values and is selected based on the performance requirements.
\end{remark}

The control input, along with the predicted future trajectory, is inferred by solving the factor graph for the prediction horizon $n$. 
Following the execution of the first control and receiving new estimates from optimizing the variable nodes, we extend the time horizon one step further and repeat the entire process.
Algorithm~\ref{alg:Planning} outlines this methodology of the proposed framework.
\begin{algorithm}[t]
\SetKwFunction{Init}{Init}
\SetKwFunction{LocalUpdate}{LocalUpdate}
\SetKwFunction{BinaryUIpdate}{BinaryUIpdate}
\SetKwProg{UP}{UnitaryUpdate}{}{}
\SetKwProg{BUP}{BinaryUIpdate}{}{}

\KwData{track boundaries $\bm{b}_{l}$, $\bm{b}_{r}$,
        centerline $\bm{p}$, number of states $N$,
        factor graph $\mathcal{G}=\left \{ \bm{\Theta}, \mathcal{F},\mathcal{E}\right\}$,
        prediction horizon $n$, convergence threshold for the optimizer $\eta$
        }
\KwResult{$\bm{u}^{*}$ and $\bm{\theta}^{*}$}
\vspace{5pt}
Initialize: Compute SDF \Comment*[r]{(Sec.~\ref{sec:results})}
\For{${j} = 0,1, ..., N-1$}{
    \tcc{factor graph for horizon $n$}
    $\prod_{m=j}^{j+n} \;f_{m}\left ( \bm{\Theta}_{m} \right )$ 
    \vspace{5pt}
    
    \While{$\Delta \bm{\Theta} \; \geq \eta$}{
    \tcc{solve factor graph}
    $\underset{\bm{\Theta} }{\arg\max} \;
            \prod_{m=j}^{j+n} \;f_{m}\left ( \bm{\Theta}_{m} \right )$ 
            
   $\bm{\Theta} \leftarrow  \bm{\Theta} +\delta\bm{\Theta}$
   }
    apply: $\bm{u}_{j}^{*}$ 
    
    obtain: $\bm{\theta}^{*}_{j:n+j}$ 
    
    update graph nodes: $\bm{\theta}_{j:n+j}$ $\leftarrow \;$  $\bm{\theta}^{*}_{j:n+j}$
    
}
\vspace{5pt}
return $\bm{\Theta}^{*}$
\caption{minCurvFG: Minimum curvature planning via factor graph for an autonomous racecar.}
\label{alg:Planning}
\end{algorithm}
\section{EVALUATION} \label{sec:results}
The proposed framework is evaluated for two exemplary racetracks.
The \ac{MPCC}-based optimization algorithm~\cite{Liniger15} is used as a benchmark for a 1:43 scale racecar.
Two solvers, \ac{HPIPM}\cite{Frison20} and the MATLAB \ac{QP} solver (quadprog), are utilized for the \ac{MPCC} benchmark.
The algorithms are evaluated by running tests for both the racetracks to note cumulative curvature, total distance, lap time, and computation time for each optimization step. 
While opting for a 1:43 scale racecar for algorithm evaluation proves to be an effective choice, capturing its performance well, we also present results for a full-scale racing car to further validate the proposed algorithm's capabilities.
All tests are conducted on a 2.1 GHz AMD Ryzen 5 PRO 3500U quad-core processor with 16GB of RAM.

\subsection{IMPLEMENTATION DETAILS} \label{sec:implDetails}
The proposed framework has been implemented using the \ac{GTSAM} library, similar to the approach employed in \ac{GPMP2}~\cite{MukadamDYDB18}. 
The algorithm is structured into two layers: the top layer is developed in MATLAB and encompasses the vehicle model, racetrack specifications, and trajectory characteristics. 
On the lower level, the factor error functions are implemented in C++ to augment the capabilities of the \ac{GTSAM} library.
The Levenberg–Marquardt algorithm from the \ac{GTSAM} is employed as the numerical optimization solver.

Following the trend observed in recent planning algorithms \cite{MukadamDYDB18, BariGW23}, the obstacle cost is $\mathbf{e}^{obs}\left ( \bm{p}_{i}\right ) = \mathbf{c}\left ( d \right )$, where, $\mathbf{c}$ is the hinge loss function such as,
\begin{equation}
\mathbf{c}\left ( d \right )= \left\{\begin{matrix}
-d+\varepsilon   & \text{if} \;\; d\leq \varepsilon  \\ 
 0&  \text{if} \;\; d>  \varepsilon 
\end{matrix}\right.,
    \label{eq:hinge}
\end{equation}
where $d$ denotes the signed distance to the closest boundary, and $\varepsilon$ represents the safety distance. 
As outlined in Section~\ref{sec:methodology}, the computations for signed distance fields are conducted offline prior to the commencement of the race.

For our experiments, we set $\varepsilon$ to 0.015cm for 1:43 scale car and 0.5m for the full-scale car. 
The selection of a smaller safety distance is motivated by the observation that in autonomous racing, minimum curvature trajectories often lie close to the boundary, enabling higher speeds.
A horizon length of $n = 40$ with a sampling time of $T_s = 20$ ms is chosen for all experiments, resulting in a prediction horizon of 0.8 seconds. 
Further details regarding the model parameters for the 1:43 and full-scale racecar, state, and control input constraint limits, and the selected values of the hyperparameters of factor nodes can be found in Appendix B.
\subsection{RESULTS FOR 1:43 SCALED CAR}
\begin{figure}[t]
    \centering
    \includegraphics[width =\linewidth]{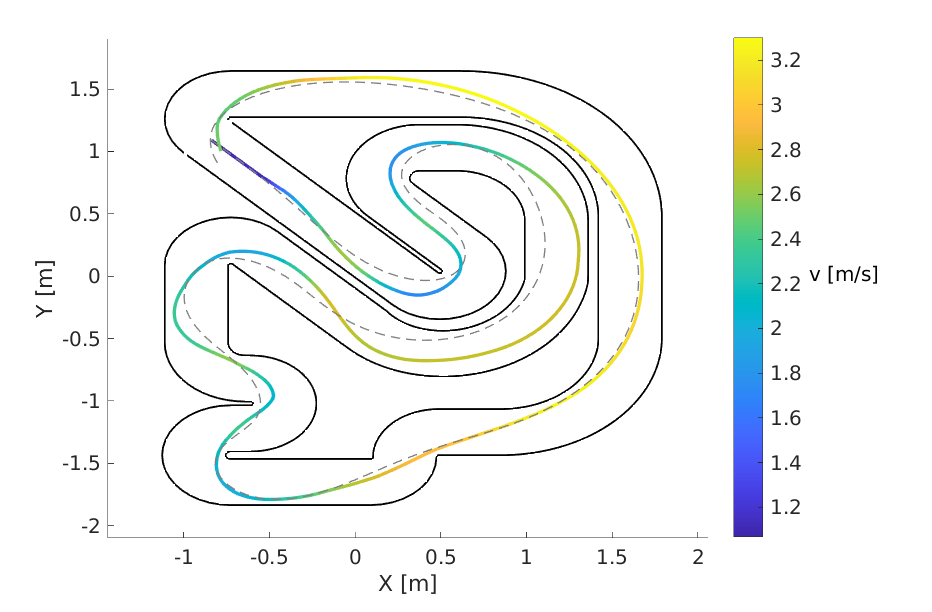}
    \caption{The trajectory generated by the proposed algorithm in case of 1:43 scaled car for one lap on Track-I.}
    \label{fig:track1}
\end{figure}

The trajectory produced by the proposed algorithm for one lap is depicted in Fig.~\ref{fig:track1}, along with the velocity profile. 
The proposed approach excels with the resulting cumulative curvature of 767.26, outperforming the benchmark. 
This improvement is attributed to the inclusion of the factor node $f^{\mathrm{curv}}$ in the factor graph, specifically designed to minimize the cumulative curvature. 
Table~\ref{tab:track1Results} provides a summary of the benchmark results obtained for Track-I.

There is a trade-off between the shortest path and the minimum curvature path.
The increase in total covered distance on the racetrack was also observed for the case with the minimum curvature path (as outlined in Table~\ref{tab:track1Results}). 
In the context of autonomous racing, the prioritization of a superior velocity profile for better lap time justifies the preference for reduced curvature, even with a marginal increase in distances. 
The impact of reduced curvature is distinctly visible in the improved velocity profile for Track-I.
In our case, the mean speed achieved is 2.47m/s, compared to 2.03m/s for the benchmark. 
Additionally, the increases in maximum speed (from 3.22m/s to 3.30m/s) are recorded.
The increase in mean speed across the racetrack ensures a better lap time of 6.93s demonstrating the effect of incorporating minimum curvature characteristic. 

While our approach demonstrates computational efficiency compared to the quadprog solver for \ac{MPCC}, the \ac{HPIPM} is more efficient in terms of providing fast solutions.
The comparatively better efficiency of the proposed approach as compared to \ac{MPCC}-quadprog solver is mainly attributed to the proposed factor graph architecture that leads to sparse non-linear least squares optimization.
However, \ac{HPIPM} solver proved to be better in terms of computational efficiency. 
We believe that the computation efficiency of the factor graph-based planning algorithms can be further improved by leveraging the parallel processing capabilities (ref. to Sec.~\ref{sec:discussion} for more details).

\begin{figure}[t]
    \centering
    \includegraphics[width =\linewidth]{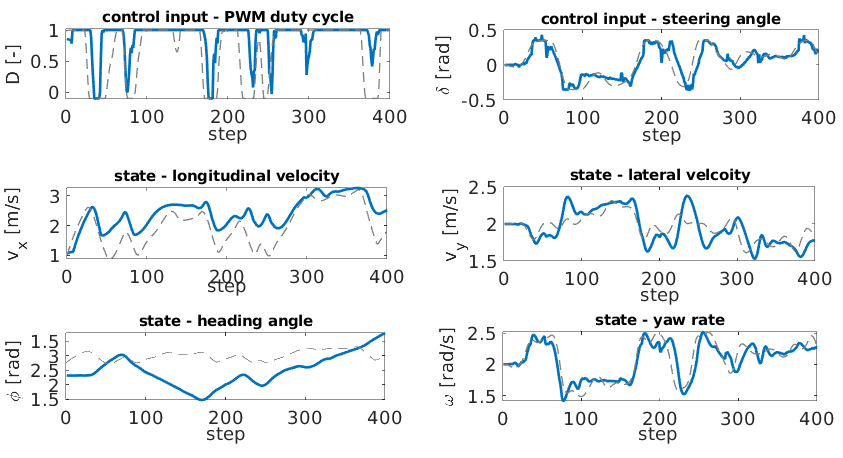}
    \caption{State and control input of 1:43 scaled car for Track-I.}
    \label{fig:track1values}
\end{figure}
\begin{table*}[th]
\centering
\caption{Track-I quantitative results of 1:43 scaled car obtained for one lap. Bold values represent the best performing results.}
\label{tab:track1Results}
\begin{tabular}{p{2.5cm} p{0.1cm} p{1cm} p{1.2cm} p{1.4cm} p{1.34cm} p{1.5cm} p{1.45cm} p{1.4cm}}
\toprule
 &
  \multicolumn{1}{r}{Cumulative Curvature} & Distance (m) & Lap Time (s)  & Mean Speed (m/s) & Max. Speed (m/s) & Mean Comp. Time (ms) & Max. Comp. Time (ms) & Min. Comp. Time (ms) \\
  \midrule
minCurvFG & \textbf{767.26} &  17.14 & \textbf{6.93} &\textbf{2.47} & \textbf{3.30} & 14.14 & 236.44 & 5.32 \\
MPCC-HPIPM~\cite{Liniger15} & 814.63 &  \textbf{15.99} & 7.87 & 2.03 & 3.22 & \textbf{3.96} & \textbf{9.90} & \textbf{2.71} \\
MPCC-quadprog~\cite{Liniger15} & 814.60 & 16.00 & 7.88 & 2.03 & 3.22 & 899.90 & 2002.70 & 497.76 \\  
  \bottomrule
\end{tabular}%
\end{table*}

Fig.~\ref{fig:track1values} illustrates the state and control input for the trajectory produced on Track-I. 
It also shows the longitudinal and lateral components of velocity for the complete racetrack.
During our experiments, we observed that the algorithm's performance is highly influenced by the hyperparameters of the factor nodes. 
Since constraints are handled softly, determining the right hyperparameters for constraint-related factors is crucial. 
Apart from the careful selection of these parameters, having a safety buffer for constraints proved to be a practical solution in case of constraint violation.

\begin{figure}[t]
    \centering
    \includegraphics[width =\linewidth]{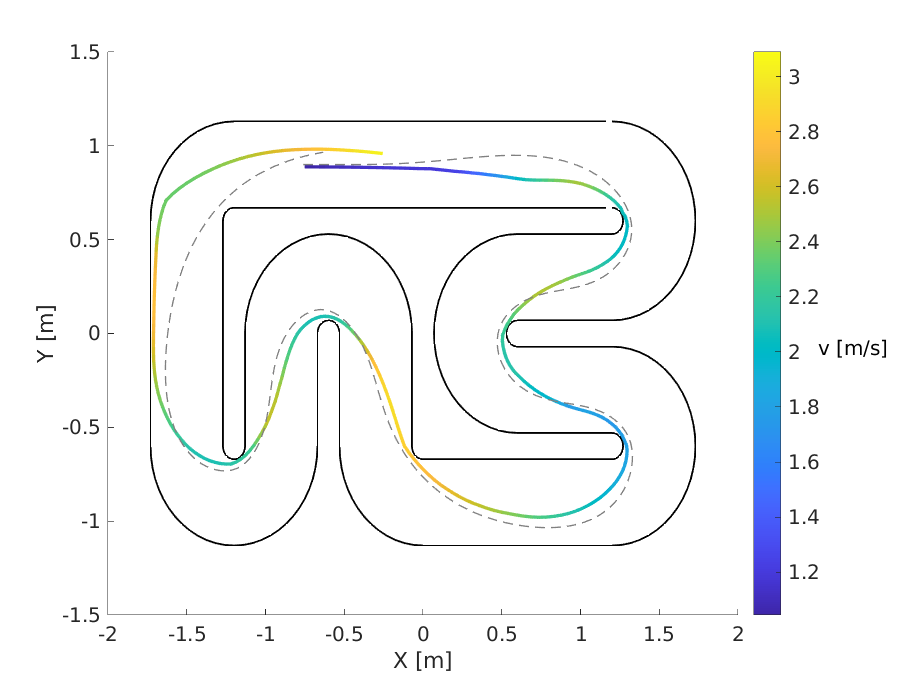}
    \caption{Output trajectory and velocity profile of 1:43 scaled car for Track-II.}
    \label{fig:track2}
\end{figure}

\begin{figure}[t]
    \centering
    \includegraphics[width =\linewidth]{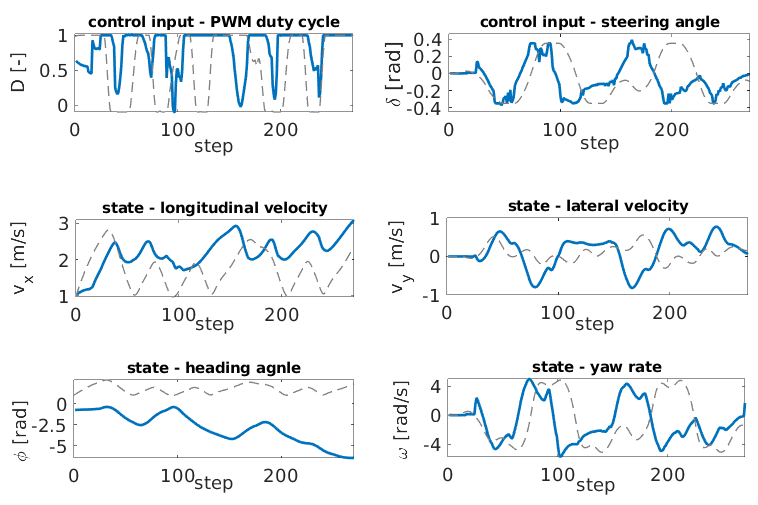}
    \caption{State and control input of 1:43 scaled car for Track-II.}
    \label{fig:track2values}
\end{figure}

Similar observations were made for Track-II, as shown in Fig.~\ref{fig:track2} and Fig.~\ref{fig:track2values}. 
A significant reduction in curvature, from 647.7 to 567.45 compared to the benchmark, is noticed, highlighting the effectiveness of our proposed algorithm.
The increase in distance covered is less in this case due to the racetrack having more sharp turns, emphasizing the algorithm's performance. 
We also observed substantial improvements in terms of speed (Table~\ref{tab:track2Results}), with a mean speed of 2.23m/s compared to 1.82m/s in the benchmark, reducing the resulting lap time to 4.92s. 
The maximum speed increased from 3.08m/s to 2.84m/s. 
Computation time results are consistent, with our proposed algorithm outperforming the quadprog solver but being less efficient than \ac{HPIPM}.

The impact of curvature minimization becomes more evident in the case of Track-II, particularly owing to its sharper corners. 
The ability to achieve a better velocity profile around these corners stems from curvature minimization, which significantly contributes to an overall improvement in the speed profile, which is a crucial factor in the context of autonomous racing competitions.
While our recorded results are only for a single lap, it is important to note that the overall planning horizon can be readily scaled to encompass multiple laps. 
In that case, the outcomes from the preceding lap can be considered as prior knowledge. 
Nonetheless, a limitation arises concerning environment perception, particularly in addressing changes in the racetrack over the laps, as the environment perception component is executed during the offline phase.
\begin{table*}[t]
\centering
\caption{Track-II quantitative results of 1:43 scaled car obtained for one lap. Bold values represent the best performing results.}
\label{tab:track2Results}
\begin{tabular}{p{2.5cm} p{0.1cm} p{1cm} p{1.2cm} p{1.4cm} p{1.34cm} p{1.5cm} p{1.45cm} p{1.4cm}}
\toprule
 &
  \multicolumn{1}{r}{Cumulative Curvature} & Distance (m) & Lap Time (s) & Mean Speed (m/s) & Max. Speed (m/s) & Mean Comp. Time (ms) & Max. Comp. Time (ms) & Min. Comp. Time (ms) \\
  \midrule
minCurvFG & \textbf{567.45} &  10.99 & \textbf{4.92} & \textbf{2.23} & \textbf{3.08}  & 23.96 & 327.38 & 6.00 \\
MPCC-HPIPM~\cite{Liniger15} & 647.75 &  \textbf{10.90} & 5.98 &  1.82 & 2.84  & \textbf{4.84} & \textbf{9.74}  & \textbf{3.03} \\
MPCC-quadprog~\cite{Liniger15} & 647.76 & \textbf{10.90} & 5.98 & 1.82 & 2.84  & 809.92 & 2036.5  & 531.88 \\  
  \bottomrule
\end{tabular}%
\end{table*}
\subsection{RESULTS FOR FULL-SCALE CAR}

\begin{figure}[t]
    \centering
    \includegraphics[width =\linewidth, height = 0.76\linewidth]{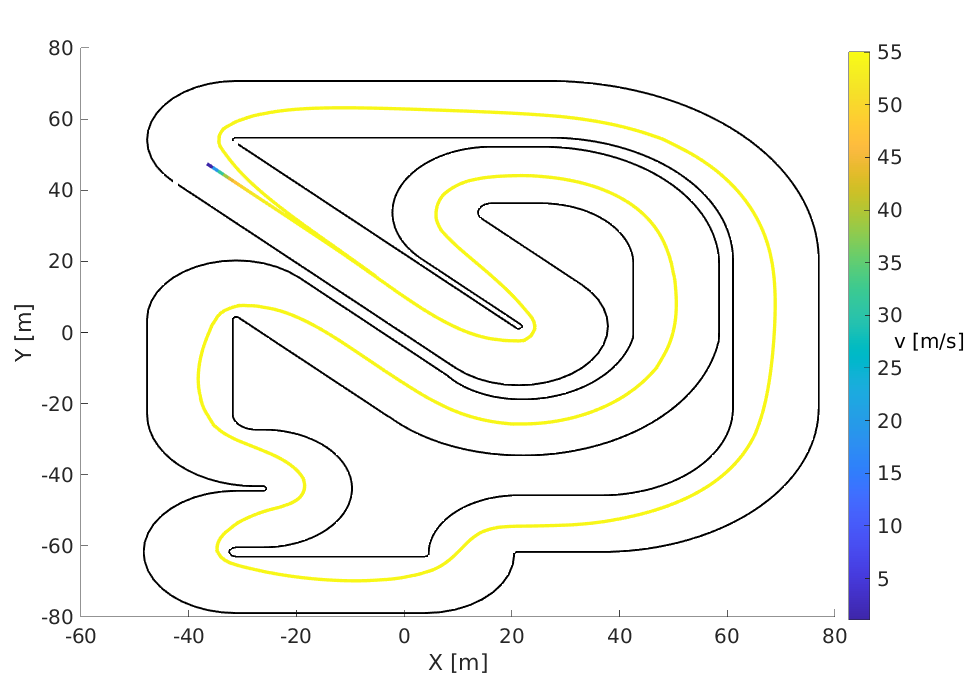}
    \caption{Full-scale car output trajectory and velocity profile for Track-I.}
    \label{fig:track1Full}
\end{figure}

\begin{figure}[t]
    \centering
    \includegraphics[width =\linewidth]{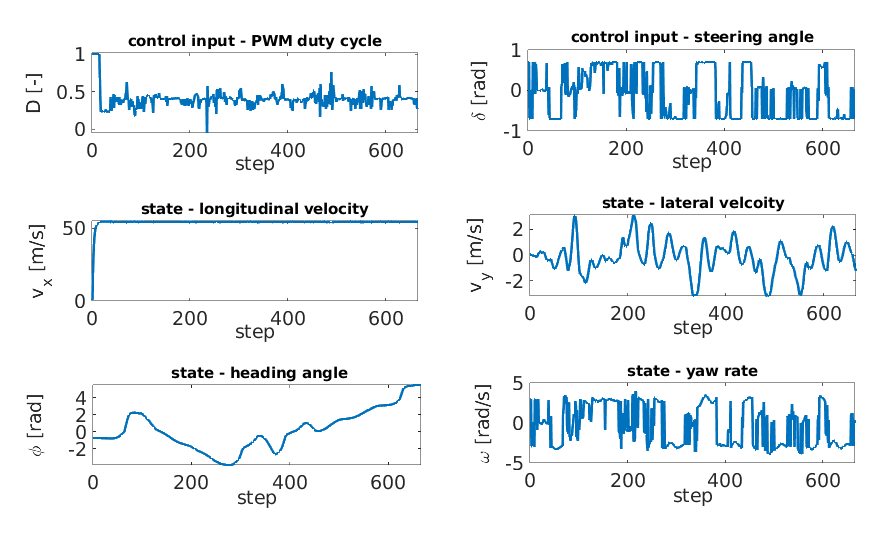}
    \caption{Full-scale car state and control input for Track-I.}
    \label{fig:track1valuesFull}
\end{figure}

\begin{figure}[t]
    \centering
    \includegraphics[width =\linewidth]{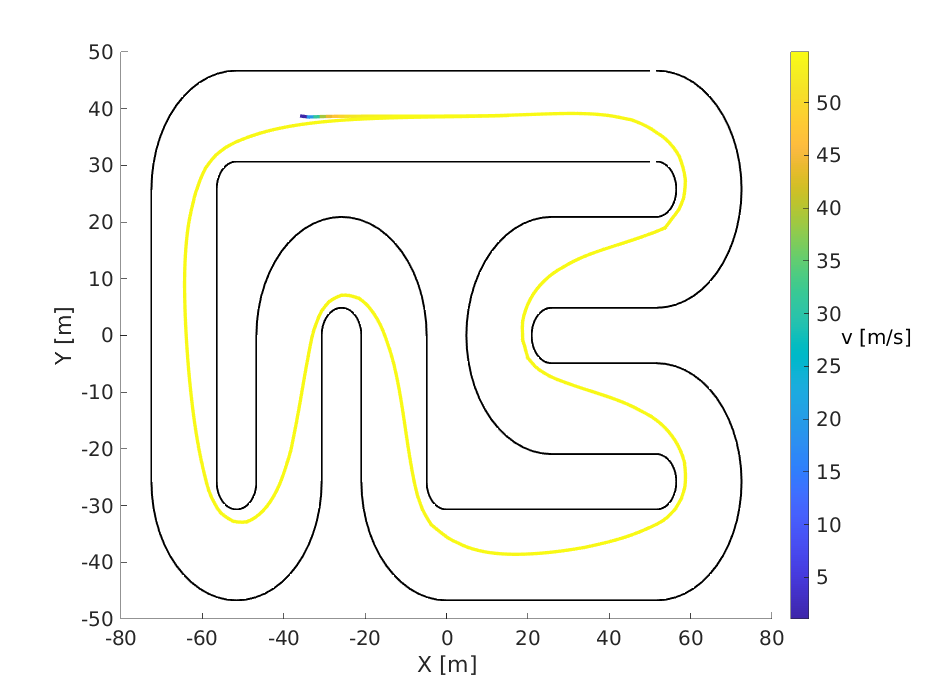}
    \caption{Full-scale car output trajectory and velocity profile for Track-II.}
    \label{fig:track2Full}
\end{figure}

\begin{figure}[t]
    \centering
    \includegraphics[width =\linewidth]{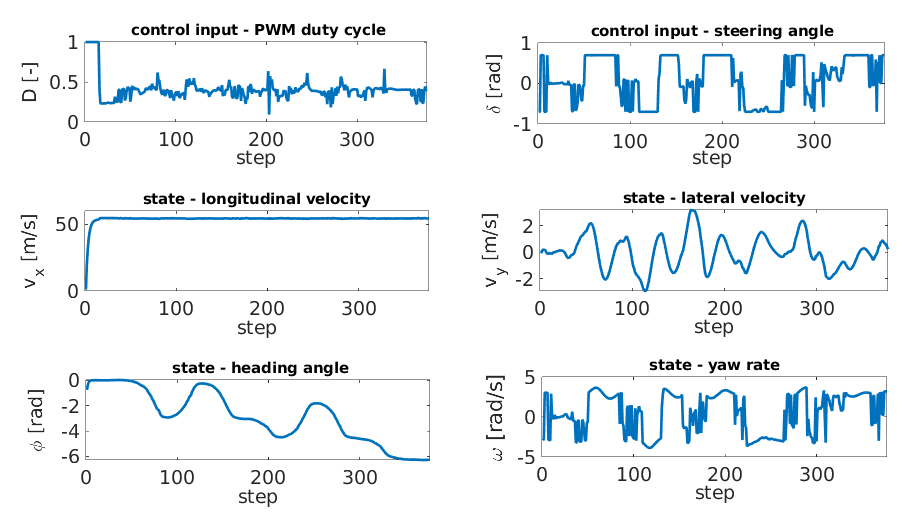}
    \caption{Full-scale car state and control input for Track-II.}
    \label{fig:track2valuesFull}
\end{figure}

The obtained results for the full-scale car are depicted in Fig.~\ref{fig:track1Full} for Track-I and Fig.~\ref{fig:track2Full} for Track-II.
For the evaluation, a maximum achievable velocity of 55.5m/s was selected. 
The findings reveal that the full-scale car consistently strives to maintain this velocity throughout the race tracks while adhering to the minimum curvature path. 
It's noteworthy that the width of the racetrack facilitates achieving and sustaining the maximum velocity throughout, leading to minimal fluctuations once this velocity is attained.
It is especially evident during cornering maneuvers. The ample space allows for smoother transitions and reduced deceleration, contributing to improved overall performance.
This observation is supported by the state and control inputs displayed in Fig.~\ref{fig:track1valuesFull} and Fig.~\ref{fig:track2valuesFull} for both racetracks, respectively. The speed profile graphs show the quick increase at the start of the race and as soon as the speed reaches the 
maximum limit, it is maintained through the racetracks by following the minimum curvature path.

A comprehensive summary of the results is presented in Table~\ref{tab:FullResults}, showcasing the effectiveness and computational efficiency of our proposed framework across both racetracks. Specifically, for Track-I, the mean speed recorded is 54.13m/s, marginally exceeding the 53.99m/s achieved on Track-II. This trend is consistent with the results obtained from the 1:43 scale racecar experiments. However, the difference in mean speed between the full-scale and scaled-down car experiments is more evident, reflecting the narrower track widths for the latter.
While the full-scale car results validate the algorithm's efficacy in real-world racing scenarios, the evaluation of the 1:43 scale car provides valuable insights into its performance under more challenging conditions.

\section{DISCUSSION} \label{sec:discussion}
The proposed approach has a strong potential as an approximation method of predictive control laws, leading to better approximation quality and computation speed, as we illustrate via simulation examples. 
However, factor graph solution via unconstrained least square optimization does not guarantee hard constraint satisfaction and feasibility. 
We noticed that, in a practical aspect, this limitation can be overcome by adding a safety buffer on top of the specified constraint limits.
A better yet trivial approach that can be adopted to overcome this limitation is having an additional layer for guaranteed constraint satisfaction purposes.
These kinds of approaches~\cite{ChenSALKPM18, KargL20} are used for approximate~\ac{MPC} solutions to guarantee constraints satisfaction and feasibility.

On the other hand, recent work~\cite{sodhi20,Bazzana22} has targeted the constrained optimization solution of factor graphs. These still rely on traditional constrained optimization approaches similar to optimization-based planning algorithms. 
A potential future research direction in terms of constraint handling in factor graphs is via message passing~\cite{YedidiaWD11, BariGW23}.
Adopting the message passing framework can also provide the opportunity to exploit the distributive properties of the factor graph in terms of parallel computing capabilities of \ac{MAP} inference that can further lead to computational efficiency.
The factor graph-based planning techniques have not yet been thoroughly studied in the field of autonomous racing. 
By exploring this avenue, efficient inference-based algorithms can be developed that can better handle the constraints and generate racing trajectories.

Formulation of autonomous racecar planning via factor graphs is a powerful approach that provides effective reasoning in designing planning objectives with better computational efficiency. 
However, the hyperparameters $\bm{\mathcal{K}}_{\left(. \right)}$ and $\bm{\Sigma}_{\left(. \right)}$ that define the stiffness of the assigned factors can strongly affect the performance in practice.
The hyperparameters need to be re-tuned if the racetrack architecture changes significantly, making the manual tuning of these parameters quite hard.
This limitation can be addressed by adapting these parameters from a trained model leveraging expert data as proposed in~\cite{BhardwajBM20}. 

Additionally, our methodology has only been tested in a single racing car scenario. 
To better understand its effectiveness, exploring its performance in a real racing setting with multiple cars and common maneuvers such as overtaking would be valuable, similar to multi-vehicle approaches presented in~\cite{WischnewskiHWL23,Thakur24}.
However, investigating such scenarios is beyond the scope of this work and remains a potential avenue for future research. 
Nevertheless, our proposed framework offers the flexibility to incorporate additional racing criteria through enhancements to the factor graph and further racing functionalities for multi-vehicles~\cite{WischnewskiHWL23,Thakur24} can be incorporated, suggesting that addressing these challenges within our framework is a promising area for future exploration.
\begin{table*}[t]
\centering
\caption{Quantitative results of full-scale car obtained for one lap.}
\label{tab:FullResults}
\begin{tabular}{p{1.3cm} p{0.1cm} p{1cm} p{1.2cm} p{1.4cm} p{1.34cm} p{1.5cm} p{1.5cm} p{1.45cm} p{1.4cm}}
\toprule
 &
  \multicolumn{1}{r}{Cumulative Curvature} & Distance (m) & Lap Time (s) &  Mean Speed (m/s) & Max. Speed (m/s) & Mean Comp. Time (ms) & Max. Comp. Time (ms) & Min. Comp. Time (ms) \\
  \midrule
Track-I & 29.82 &  740.97 & 13.68 & 54.14 & 55.01 & 35.08 & 239.28 & 10.34 \\
Track-II & 16.79 &  516.04 & 9.56  & 53.99 & 54.7 & 29.49  &  145.35 & 4.44 \\ 
  \bottomrule
\end{tabular}%
\end{table*}


\section{CONCLUSION} \label{sec:conclusions}
This work proposes an alternate perspective on the autonomous racecar planning problem by utilizing probabilistic inference-based problem formulation.
The proposed framework decomposes the racecar planning problem by exploiting the inherent structure among random variables and formulates the planning objectives as a joint probability distribution using a factor graph.
The integration of the global planning objective of curvature minimization into the local planning module via a novel factor node design highlights the utility of the factor graph-based formulation in modeling planning problems. 
This structured representation of the planning problem is computationally efficient and produces results that are comparatively better than the optimization-based benchmark.
This work has opened a way forward to exploit the distributive properties of factor graphs by utilizing message passing-based solution techniques.
Parallel processing and storage during message passing among graph nodes and different strategies for message scheduling can further improve the computational efficiency that is crucial in competitive autonomous racing scenarios.
\section*{APPENDICES}
\subsection*{A. PROOF OF PROPOSITION~\ref{prop:angle}}\label{app:proof}
\begin{figure}[t]
    \centering
    \includegraphics[width =\linewidth]{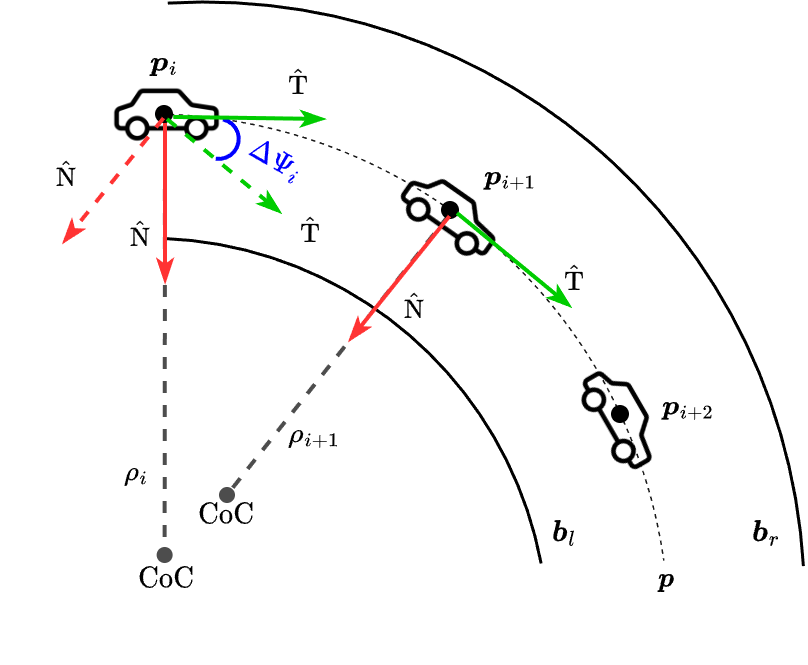}
    \caption{The figure shows an example of a racetrack represented in the N-T coordinate system. The arrows $\hat{\mathrm{N}}$ and $\hat{\mathrm{T}}$ indicate the normal-tangent unit vectors projected from  $\bm{p}_{i+1}$ onto $\bm{p}_{i}$, which illustrate the formation of tangential rotation angle $\Delta \Psi_{i}$.}
    \label{fig:tangentangle}
\end{figure}

Assuming curvilinear motion of the racecar, we do not consider a fixed \acl{CoC}. Instead, the origin is located on the individual states along the curve and represented on an \ac{N-T} coordinate system, as shown in Fig.~\ref{fig:tangentangle}. 
The radius of curvature, denoted by $\rho$, is defined as the perpendicular distance between the curve and the \ac{CoC}. 
The unit vectors $\hat{\mathrm{N}}$ and $\hat{\mathrm{T}}$ are always oriented in the positive direction of the N-axis and T-axis, respectively.
The radius $\rho$ determines the amount of curvature $k$ of an arc. Curvature $k$ and radius $\rho$ have an inverse relationship. Therefore, the greater the curvature $k$, the smaller the radius $\rho$ of an arc, and vice versa.

The rotation of tangents between two states along the curve characterizes the curvature such that, as the $\Delta \Psi \rightarrow  0$, the radius of curvature $\rho  \rightarrow  \infty $. 
The relation between the $\Delta \Psi_{i}$ and the curvature $k_{i}$ can be observed from Fig. \ref{fig:tangentangle}. 
Since the unit vectors $\hat{T}$ and $\hat{N}$ are perpendicular at any point of the arc, the angle $\Delta \Psi_{i}$ along which the tangent rotates is also the same angle along which the normal vector rotates. 
If the time interval $\Delta t$ is very small, the two unit vectors meet at the same \ac{CoC}. 
By projecting the point $\bm{p}_{i+1}$ onto $\bm{p}_{i}$, it can be observed (from Fig. \ref{fig:tangentangle}) that by reducing the  $\Delta \Psi_{i}$, both the points will lie on the same straight line resulting in $\rho  \rightarrow  \infty $ and the curvature $k_{i}$ that reduces the overall quadratic curvature $k$.
\begin{table*}[t]
\centering
\caption{The experimental parameter values for the covariance matrices of each factor node in the factor graph. }
\label{append:simParams}
\begin{tabular}{ccccccccc}
\toprule
  Parameter &$\sigma_{\mathrm{s | g}}$ & $\sigma_{\mathrm{ref}}$ &  $\sigma_{\mathrm{vel}}$ & $\sigma_{\mathrm{rlim}}$ & $\sigma_{\mathrm{ulim}}$ & $\sigma_{\mathrm{obs}}$ & $\sigma_{\mathrm{sys}}$ & $\sigma_{\mathrm{curv}}$ \\
  \midrule
Value [1:43 scale] & 8e-4 &  5.7e-2 & 5.5e-2 & 1e-3 & 5e-6 & 1e-4 & 1e-5 & 1e-2 | 3.1e-2 \\
\midrule
Value [Full scale] & 8e-4 &  5.7e-2 & 5.5e-2 & 1e-3 & 5e-6 & 1e-5 & 1e-3 & 5e-3 \\
  \bottomrule
\end{tabular}%
\end{table*}

\begin{table}[ht]
\centering
\caption{Model parameters.}
\label{append:modelParams}
\begin{tabular}{@{}lcc}
\toprule
 &
  \multicolumn{1}{l}{Value [1:43 scale] } & {Value [Full scale]} \\
  \midrule
$m$ [mass]-kg &  0.041   & 1573 \\
$I_z$ [inertia]-$\mathrm{kg.m^{2}}$ &  27.8e-6  & 2873\\
$l_{f}$ [\ac{CoG} to front axle]-m & 0.029 & 1.35 \\
$l_{b}$ [\ac{CoG} to rear axle]-m & 0.033 & 1.35 \\ 
$L$ [length of racecar]-m & 0.12 &  5\\ 
$W$ [width of racecar]-m & 0.06 &  2.5\\ 
$C_r0$ [rolling resistance] & 0.0518 &  120\\ 
$C_m1$ [tire specific constant] & 0.287 &  17303\\ 
$C_m2$ [tire specific constant] & 0.0545 & 175\\ 
$C_d$ [tire specific constant] & 0.00035 &  0.535\\ 
$B_b$ [tire specific constant] & 3.3852  & 13\\ 
$C_b$ [tire specific constant] & 1.2691  & 2\\ 
$D_b$ [tire specific constant] & 0.1737 &  9258.6\\ 
$B_f$ [tire specific constant] & 2.579 &  13\\ 
$C_f$ [tire specific constant] & 1.2 & 2\\ 
$D_f$ [tire specific constant] & 0.192 &  9258.6\\ 
  \bottomrule
\end{tabular}%
\end{table}
\begin{table}[th]
\centering
\caption{Limits on the state and control.}
\label{append:limits}
\begin{tabular}{@{}lcc}
\toprule
 &
  \multicolumn{1}{l}{Range [1:43 scale] }& {Range [Full scale]}  \\
  \midrule
$v_{x}$ [m/s]  & [-0.1 4.0] & [-0.1 55.5]\\
$v_{y}$ [m/s] & [-2.5 2.5] &  [-2.5 2.5]\\ 
$ \varphi$ [rad] & [-10.0 10.0] & [-10.0 10.0] \\ 
$ \omega$ [rad/s] & [-7.0 7.0] &  [-7.0 7.0]\\ 
$\delta$ [rad]  & [-0.4 0.4] &   [-0.7 7.0]\\ 
$d$  & [-0.1 1.0]  & [-0.1 1.0]\\ 
  \bottomrule
\end{tabular}
\end{table}
\subsection*{B. MODEL AND SIMULATION PARAMETERS}\label{app:params}
The hyperparameter matrices of the prior factors $\bm{\mathcal{K}}$ and likelihood factors $\bm{\Sigma}$ are defined as $\sigma^{2}\mathbf{I}$, where,  $\sigma$ is the variance for the respective factor constraint, indicating how "tight" the constraint is.
The values chosen for each factor attached to the factor graph are detailed in Table~\ref{append:simParams} for both the 1:43 and full-scale racecars.
Note that for the minimum curvature factor $f^{\mathrm{curv}}$ in case of 1:43 racecar, the value is set to 1e-2 for Track-I and 3.1e-2 for Track-II.
The racecar model parameters are presented in Table~\ref{append:modelParams} (adopted from~\cite{Liniger15}). 
The limitations on state and control are depicted in Table~\ref{append:limits}.

\section*{ACKNOWLEDGMENT}
The authors thank Michael Fink, and Marion Leibold for valuable discussions.
\section*{REFERENCES}

\renewcommand{\refname}{} 
\bibliography{IEEEabrv,mybibfile}
\bibliographystyle{IEEEtran}
\begin{IEEEbiography}[{\includegraphics[width=1in,height=1.25in,clip,keepaspectratio]{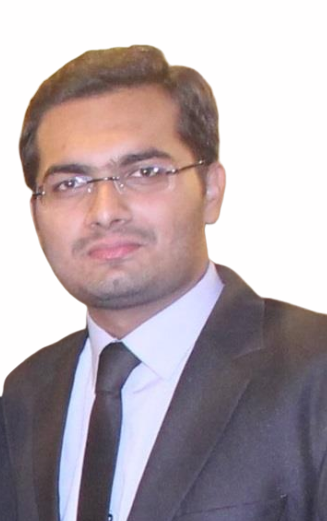}}]{Salman Bari } received a B.Sc. degree in Electrical Engineering from the University of Engineering and Technology, Lahore, Pakistan, in 2014 and an M.Sc. degree in Mechatronics Engineering from Air University, Islamabad, Pakistan, in 2018.
He is currently a Research Associate and a Ph.D. candidate at the Chair of Automatic Control Engineering (LSR), TUM School of Computation, Information and Technology, Technical University of Munich, Germany. 
His research interests include robot motion planning and probabilistic machine learning.
\end{IEEEbiography}

\begin{IEEEbiography}[{\includegraphics[width=1in,height=1.25in,clip,keepaspectratio]{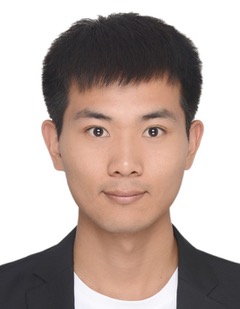}}]{Xiagong Wang } received the B.Sc. degree in Electrical and  Electronic Engineering from the University of Duisburg Essen in 2021 and M.Sc.
degree in Electrical Engineering and Information Technology from TUM School of Computation, Information and Technology, Technical University of Munich, Germany, in 2023. 
His research interests include graphical models for planning and their applications in Autonomous Racing. 
\end{IEEEbiography}

\begin{IEEEbiography}[{\includegraphics[width=1in,height=1.25in,clip,keepaspectratio]{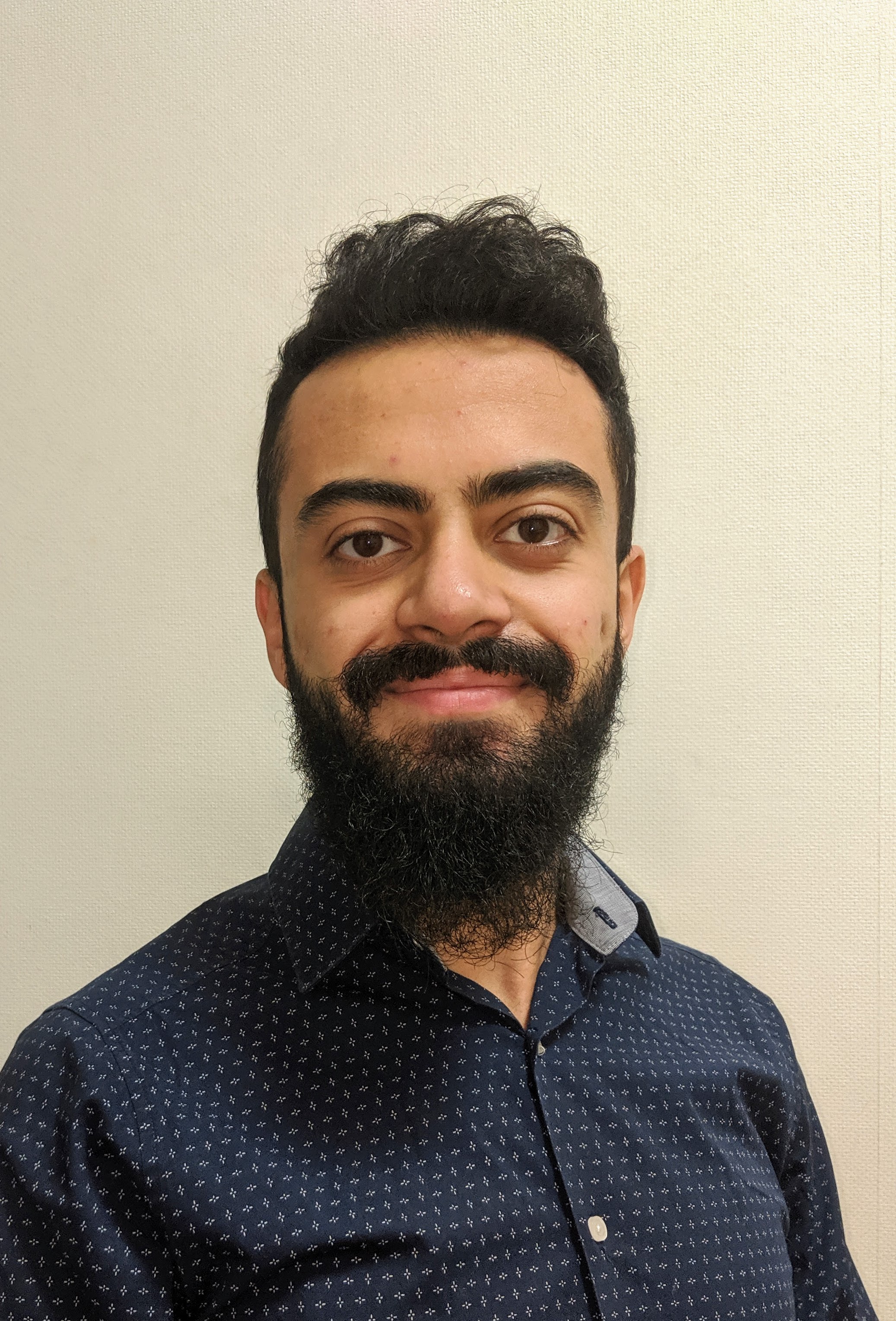}}]{Ahmad Schoha Haidari } received the B.Sc. degree in Electrical Engineering and Computer Science in 2018 and the M.Sc. degree in 2022 from the Technical University of Munich, Germany. 
Since finishing his degree, he has been working at ARRK Engineering GmbH, Munich, Germany, as a Test Engineer with the main focus on infotainment system of vehicles from the BMW Group.
Since 10/2023, he is also a Senior Engineer at ARRK Engineering.

\end{IEEEbiography}

\begin{IEEEbiography}[{\includegraphics[width=1in,height=1.25in,clip,keepaspectratio]{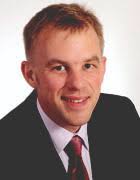}}]{Dirk Wollherr } received the Dipl.-Ing. degree and Dr.-Ing., and Habilitation degrees in electrical engineering from Technical University Munich, Germany, in 2000, 2005, and 2013, respectively.
He is currently a Senior Researcher in robotics, control and cognitive systems at the Chair of Automatic Control Engineering, TUM School of Computation, Information and Technology, Technical University of Munich, Germany. His research interests include automatic control, robotics,
autonomous mobile robots, human-robot interaction, and socially aware collaboration and joint action.
\end{IEEEbiography}
\end{document}